\documentclass{article}

\usepackage{hyperref}

\usepackage[accepted]{icml2025}

\usepackage{microtype}
\usepackage{graphicx}
\usepackage{subfigure}
\usepackage{booktabs} 
\usepackage{multirow}
\usepackage{xcolor}
\usepackage{afterpage}
\usepackage{xspace}
\usepackage{caption}
\usepackage{placeins}
\usepackage{adjustbox}

\usepackage{amsmath}
\usepackage{amssymb}
\usepackage{mathtools}
\usepackage{amsthm}

\usepackage{pifont}
\usepackage{makecell}

\usepackage[capitalize,noabbrev]{cleveref}

\newcommand{\cmark}{\ding{51}}%
\newcommand{\xmark}{\ding{55}}%

\newcommand{\weburl}{\url{https://showlab.github.io/Impossible-Videos/}\xspace}

\usepackage[symbol]{footmisc}

\usepackage[toc,page]{appendix}
\usepackage{titletoc}

\theoremstyle{plain}

\theoremstyle{definition}

\theoremstyle{remark}

\usepackage[textsize=tiny]{todonotes}

\usepackage{colortbl}
\definecolor{mygray}{gray}{0.95}

\icmltitlerunning{Impossible Videos}

\begin{document}

\twocolumn[
\icmltitle{Impossible Videos}

\icmlsetsymbol{equal}{*}

\begin{icmlauthorlist}
\icmlauthor{Zechen Bai}{equal}
\icmlauthor{Hai Ci}{equal}
\icmlauthor{Mike Zheng Shou}{}
\end{icmlauthorlist}

\icmlcorrespondingauthor{Mike Zheng Shou}{mike.zheng.shou@gmail.com}

\begin{center}
Show Lab, National University of Singapore
\end{center}

\icmlkeywords{Benchmark, Video Understanding, Video Generation}

\begin{center}
    \centering
    \weburl
    \captionsetup{type=figure}
    \includegraphics{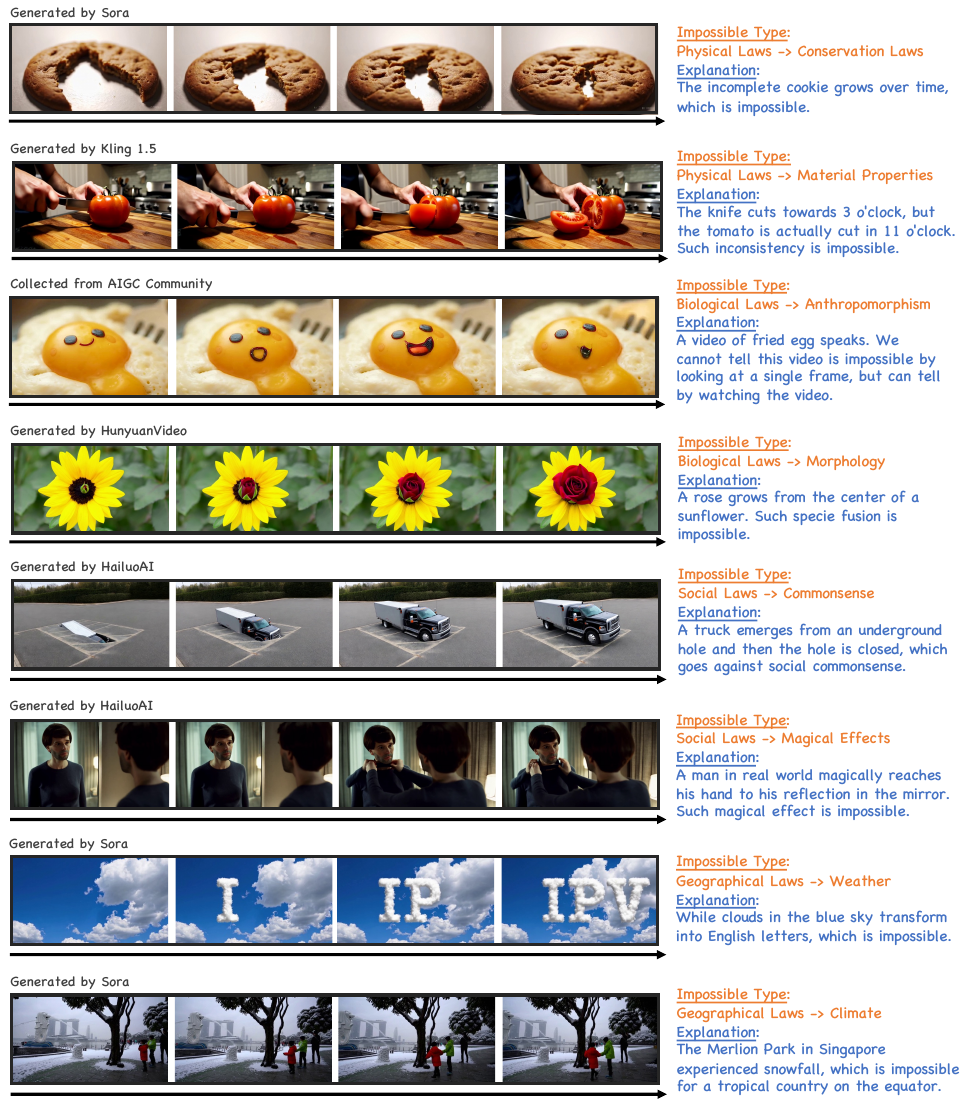}
    \vspace{-1em}
    \captionof{figure}{
        \textbf{Impossible Video Examples with Impossible Type and Explanation.}
    }
    \label{fig:teaser_vids}
\end{center}

\vskip 0.2in
]



\printAffiliationsAndNotice{\icmlEqualContribution} 

\begin{abstract}
Synthetic videos nowadays is widely used to complement data scarcity and diversity of real-world videos.
Current synthetic datasets primarily replicate real-world scenarios, leaving impossible, counterfactual and anti-reality video concepts underexplored.
This work aims to answer two questions:
1) Can today's video generation models effectively follow prompts to create impossible video content?
2) Are today's video understanding models good enough for understanding impossible videos?
To this end, we introduce \textsc{IPV-Bench}, a novel benchmark designed to evaluate and foster progress in video understanding and generation.
\textsc{IPV-Bench} is underpinned by a comprehensive taxonomy, encompassing 4 domains, 14 categories.
It features diverse scenes that defy physical, biological, geographical, or social laws.
Based on the taxonomy, a prompt suite is constructed to evaluate video generation models, challenging their prompt following and creativity capabilities.
In addition, a video benchmark is curated to assess Video-LLMs on their ability of understanding impossible videos, which particularly requires reasoning on temporal dynamics and world knowledge.
Comprehensive evaluations reveal limitations and insights for future directions of video models, paving the way for next-generation video models.
\end{abstract}

\section{Introduction}
\label{sec:introduction}

Video data has been a long-standing focus in the research community, offering the potential to capture richer and more structured information compared to text~\cite{yang2024video,liu2024world}.
Over time, this domain has expanded from early tasks like action recognition~\cite{action_recognition_survey} to more advanced applications such as video captioning~\cite{caption_survey}, question answering, video generation~\cite{video_bench}, and video-based world modeling~\cite{liu2024world}, showcasing its versatility and growing importance in advancing AI capabilities.

Although video data seems to be abundant on the Internet, it suffers from issues of low quality and diversity, etc.
Recent efforts have been trying to alleviate these issues by generating videos either from neural generation models or simulation engines~\cite{agarwal2025cosmos}.
However, the primary goal is still to replicate real-world scenes in a controlled way, which severely limits the broader imagination space and applications.
In this work, we introduce the concept of \textit{impossible videos}, which particularly captures counterfactual and anti-reality scenes that are \textit{impossible} in real world.
We argue that impossible videos can form an effective testbed to assess video models.
As an \textit{out-of-real-world-distribution} data, it requires the model to not simply \textit{memorize} real-world data and \textit{retrieve} similar information based on the input, but to genuinely \textit{learn} from real-world data and \textit{reason} upon the input.

Currently, mainstream video models can be categorized into two majoy categories based on tasks: understanding and generation~\cite{xie2024show,wu2024janus}.
Thus, this work aims to probe the boundary of the two types of models by answering the following questions.

First, \textbf{can today's video generation models effectively follow text prompts to create impossible video content?}
Recent research focus of video generation has been evolving from fundamental video quality~\cite{video_bench} (e.g., aesthetic quality, motion smoothness) to advanced semantic quality (e.g., physical laws, subject consistency)~\cite{bansal2024videophy}.
State-of-the-art models have been positioned as a world simulator~\cite{sora}.
Ideally, the model should be able to generate either physically coherent or anti-reality videos with detailed control of the text prompt, enabling wider range of applications, such as filming, advertising, etc.
Prompting the model to generate impossible videos challenges the model to ``break the rule'' yet faithfully following the prompt.

Second, \textbf{are today's video understanding models good enough for understanding impossible videos?}
Advanced video understanding models, especially video large language models (Video-LLMs), are mostly trained on large-scale video datasets, demonstrating remarkable performance on popular benchmarks with real-world videos.
Impossible videos pose specific challenges on reasoning temporal dynamics and world knowledge.

\begin{table*}[ht]
\centering
\caption{
Comparison of \textsc{IPV-Bench} and Existing Benchmarks.}
\resizebox{0.9\linewidth}{!}{
\begin{tabular}{l|ccc|ccc|ccc}

\toprule

\multirow{2}{*}{\textbf{Benchmark}} & \multicolumn{3}{c|}{\textbf{Tasks}} & \multicolumn{3}{c|}{\textbf{Video Data}} & \multicolumn{3}{c}{\textbf{Text Data}} \\ 

\cmidrule(lr){2-4} \cmidrule(lr){5-7} \cmidrule(lr){8-10}
      & \makecell{AIGC\\Detection} & \makecell{Video\\Understanding} & \makecell{Video\\Generation}  
      & \makecell{Real World\\Videos} & \makecell{Generated\\Videos} & \makecell{Impossible\\Videos}
      & \makecell{Text\\Prompts} & \makecell{Text\\Descriptions} & \makecell{Impossible\\Text}\\
\midrule
GenVideo~\cite{gen_video}       & \cmark & \xmark & \xmark & \cmark & \cmark & \xmark & \cmark & \xmark & \xmark \\
GenVidBench~\cite{genvidbench}    & \cmark & \xmark & \xmark & \cmark & \cmark & \xmark & \cmark & \xmark & \xmark \\
LOKI~\cite{ye2024loki}           & \cmark & \cmark & \xmark & \cmark & \cmark & \xmark & \xmark & \cmark & \xmark \\
\hline
VBench~\cite{huang2024vbench}         & \xmark & \xmark & \cmark & \xmark & \xmark & \xmark & \cmark & \xmark & \xmark \\
VideoPhy~\cite{bansal2024videophy}       & \xmark & \xmark & \cmark & \xmark & \xmark & \xmark & \cmark & \xmark & \xmark \\
PhyGenBench~\cite{phygenbench}    & \xmark & \xmark & \cmark & \xmark & \xmark & \xmark & \cmark & \xmark & \xmark \\
\hline
SEED-Bench~\cite{seed_bench}     & \xmark & \cmark & \xmark & \cmark & \xmark & \xmark & \xmark & \cmark & \xmark \\
Video-Bench~\cite{video_bench}    & \xmark & \cmark & \xmark & \cmark & \xmark & \xmark & \xmark & \cmark & \xmark \\
MV-Bench~\cite{li2024mvbench}       & \xmark & \cmark & \xmark & \cmark & \xmark & \xmark & \xmark & \cmark & \xmark \\
TempCompass~\cite{liu2024tempcompass}    & \xmark & \cmark & \xmark & \cmark & \xmark & \xmark & \xmark & \cmark & \xmark \\
\hline
\textbf{IPV-Bench (Ours)}      & \cmark & \cmark & \cmark & \cmark & \cmark & \cmark & \cmark & \cmark & \cmark \\

\bottomrule

\end{tabular}
}
\vspace{-10pt}
\label{tab:bench_compare}
\end{table*}

To achieve these goals, we propose a benchmark, \textsc{IPV-Bench}, focusing on \textbf{I}m\textbf{P}ossible \textbf{V}ideos.
To the best of our knowledge, this is the first work focusing on this topic.
We start with building a comprehensive taxonomy covering diverse aspects of impossible videos, including scenes violating physical laws, biological laws, geographical laws, and social laws.
Based on the taxonomy, we construct \textsc{IPV-Txt}, a prompt suite that consists of 260 text prompts;
\textsc{IPV-Vid}, a video set with 902 high-quality videos.
Both the text prompts and videos are organized following the structure of the taxonomy, describing or displaying impossible scenes with particular consideration on temporal dynamics.
Examples of impossible videos are shown in Fig.~\ref{fig:teaser_vids}.

Based on this benchmark, we conduct comprehensive evaluations for mainstream video understanding models and generation models, suggesting that most video models fall short on impossible videos.
Extensive analysis further reveals several insights of certain limitation and future direction.
We will make the data public to inspire future research.
In summary, our contributions includes:

- To the best of our knowledge, this is the first work to identify and investigate impossible videos, which explores a critical yet absent space for advanced video understanding and generation research.
To this end, we construct a benchmark \textsc{IPV-Bench}.

- \textsc{IPV-Bench} is underpinned by a comprehensive taxonomy.
Based on the taxonomy, a prompt suite and a set of high-quality videos is collected and carefully annotated, supporting downstream evaluations.

- Based on \textsc{IPV-Bench}, we conduct extensive evaluation and analysis on mainstream video understanding and generation models, unveiling current limitations and revealing future directions.

\begin{figure*}[ht!]
    \centering
    \includegraphics[width=0.95\linewidth]{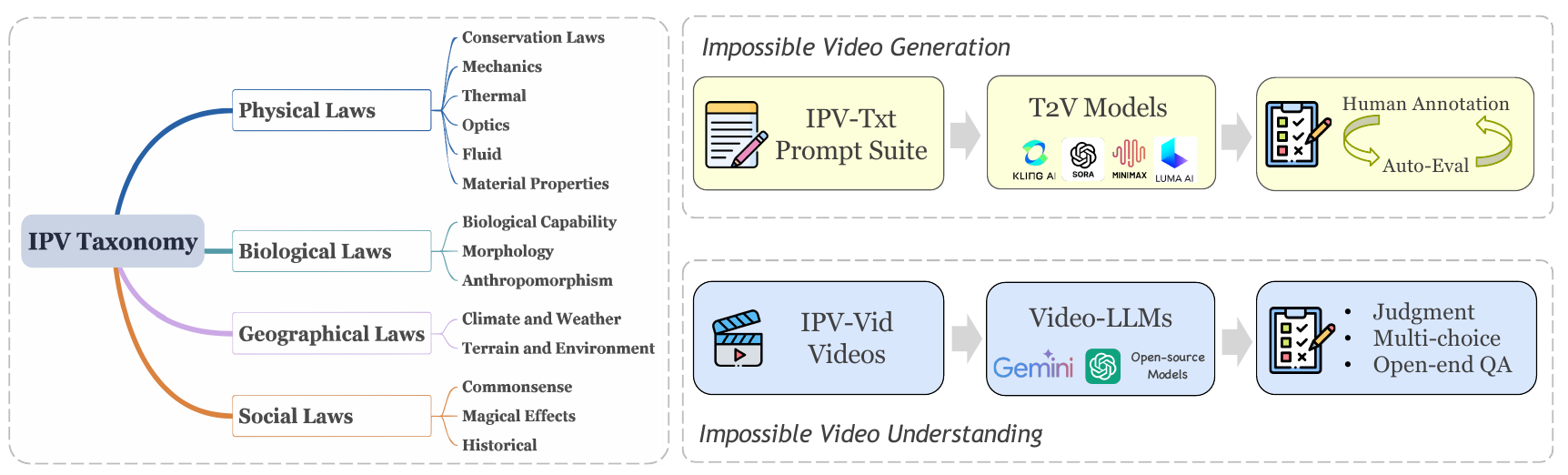}
    \vspace{-10pt}
    \caption{
    \textbf{Overview of the \textsc{IPV-Bench} Benchmark.}
    \textsc{IPV-Bench} is structured with a comprehensive taxonomy, enabling the creation of a diverse prompt suite (\textsc{IPV-Txt}) and a high-quality video dataset (\textsc{IPV-Vid}). These components facilitate the evaluation of popular video generation and understanding models.
    }
    \label{fig:main_fig}
    \vspace{-13pt}
\end{figure*}

\vspace{-5pt}
\section{Related Work}
\label{sec:related_work}
\vspace{-5pt}

This work evaluates video models across two key domains: video understanding and video generation.
Tab.~\ref{tab:bench_compare} presents a comprehensive comparison of \textsc{IPV-Bench} and existing benchmarks, highlighting their relationships and key differences.
We then outline the primary objectives of existing models and benchmarks in each domain.

\noindent \textbf{Video Understanding} remains a fundamental challenge in computer vision, encompassing tasks such as action recognition~\cite{action_recognition_survey}, object localization~\cite{vid_obj_loc,video_lisa}, tracking~\cite{ocmot}, temporal grounding~\cite{lin2023univtg}, captioning~\cite{show_recall,bai2021explain}, and, more recently, AI-Generated video detection~\cite{ye2024loki}.
Video Large Language Models (Video-LLMs)~\cite{video_llm_survey}, powered by Large Language Models (LLMs)~\cite{llm_survey}, leverage language as a universal interface to facilitate a wide range of video-related tasks.
Most open-source Video-LLMs extend from multimodal large language models (MLLMs) originally designed for images, such as LLaVA~\cite{llava}.
Some closed-source models, such as GPT-4o~\cite{gpt4o}, although initially designed for images, can also function as video models by processing multiple frames as input.
Popular benchmarks such as VideoMME~\cite{video_mme}, MV-Bench~\cite{li2024mvbench}, assess models on a range of tasks (e.g., multiple-choice and open-ended questions) and scenarios (e.g., daily life, sports, and films).
These datasets primarily consist of publicly available videos sourced from the internet.
To our knowledge, no existing benchmark explicitly includes a dedicated set of impossible or counterfactual videos for evaluation, which are crucial for assessing model generalization, robustness and reasoning abilities.

\textbf{Video Generation} has garnered significant attention in both academia and industry, with text-to-video generation serving as a foundational task~\cite{tune_a_vid}.
Notable open-source models include Stable Video Diffusion~\cite{svd}, CogVidX~\cite{yang2024cogvideox}, Open-Sora~\cite{open_sora}, Show-1~\cite{show-1}, and HunyuanVideo~\cite{kong2024hunyuanvideo}, among others. Proprietary models include Kling~\cite{klingai}, Sora~\cite{sora}, and Hailuo~\cite{hailuoai}, among others.
One of the primary challenges in video generation is achieving high visual quality, including factors such as resolution, realism, and temporal consistency~\cite{huang2024vbench}.
To address this, benchmarks such as VBench~\cite{video_bench} comprehensively evaluate various aspects of visual quality.
However, with advancements in video generation models, research focus has shifted towards ensuring semantic coherence, particularly in maintaining adherence to physical laws.
Recent benchmarks such as PhyGenBench~\cite{phygenbench} and VideoPhy~\cite{bansal2024videophy} have emerged to evaluate models' ability to generate physically plausible videos.
Beyond physics-constrained generation, an often-overlooked aspect is physics-defying content, or more broadly, the generation of impossible scenes, which plays a significant role in creative domains such as film and advertising.
Creativity is tangentially related to this challenge; however, it is only briefly considered in some comprehensive benchmarks~\cite{vid_gen_eval} and has yet to be systematically studied.

\vspace{-5pt}
\section{\textsc{IPV-Bench}}
\label{sec:benchm}
\vspace{-5pt}

We first develop a taxonomy that systematically categorizes various types of impossible scenes.
This taxonomy serves as the foundation for two critical components of the benchmark:
1) \textsc{IPV-Txt}, a suite of high-quality text prompts describing impossible scenes that cannot occur in the real world.
2) \textsc{IPV-Vid}, a curated collection of high-quality videos depicting impossible scenes, each with corresponding annotations.
An overview of the taxonomy and the roles of its components is presented in Fig.~\ref{fig:main_fig}.

\subsection{IPV Taxonomy and Prompt Suite}

\paragraph{Overview.}
As illustrated in Fig.~\ref{fig:main_fig}, the taxonomy is structured around four major categories: \textit{``Physical Laws"}, \textit{``Biological Laws"}, \textit{``Geographical Laws"}, and \textit{``Social Laws"}.
Each category is divided into multiple subcategories, which are further subdivided, forming a detailed hierarchical structure.
Building upon this hierarchy, our \textsc{IPV-Txt} benchmark incorporates 260 high-quality text prompts that describe various counterfactual and other unusual scenarios.

\begin{enumerate}
    \vspace{-12pt}
    \item 
    \textit{``Physical Laws"} covers 6 common laws: Mechanics, Thermal, Optics, Fluid, Material Properties, and Conservation Laws.
    This categorization considers most physical phenomenons.
    We instantiate text prompts to explicitly describe a scene defying a specific physical law.
    For example, ``\textit{A person pours milk into a glass cup half filled with milk, but the amount of milk in the glass cup does not change at all}" describes a video that violates the law of conservation of mass.

    \vspace{-10pt}
    \item 
    \textit{``Biological Laws"} categorizes 3 sub-categories:
    1) Biological Capability covers scenes that exceeds human or animals capabilities. For example, ``\textit{A person flapped his arms like wings and successfully flew into the sky.}"; 
    2) Morphology consider impossible body composition. E.g., ``\textit{A horse walks on the grassland, gradually growing from four legs to eight legs}";
    and 3) Anthropomorphism includes non-living objects exhibit anthropomorphic behavior, e.g., ``\textit{A fried egg with a face on it is opening its mouth and speaking something}".

    \vspace{-10pt}
    \item
    \textit{``Geographical Laws"} considers the impossible phenomenons in natural environment, including Climate and Weather anomalies, and Terrain and Environmental anomalies.
    For example, ``\textit{A mountain flattens into a perfect plateau, leaving all its trees and snow intact on the new flat surface}".

    \vspace{-10pt}
    \item
    \textit{``Social Laws"} includes 3 types of counterfactual phenomenons violating social laws:
    1) Commonsense defines unusual scenes that violates our daily routine or customs. For example, ``\textit{The programmer in front of the computer suddenly started eating the keyboard and quickly ate half of it}".
    2) Magical Effects emphasize creative content and effects that cannot be easily interpreted by certain scientific law. For example, ``\textit{e.g., A hand turns on a flashlight and shines it on a glass cup, and the cup immediately breaks into pieces.}"
    3) Historical anomalies highlight a interesting type of scene that displays impossible combinations of human-object or human-human across different historical periods. For example,
    ``\textit{Albert Einstein and Donald Trump are shaking hands at the White House}".

\vspace{-12pt}
\end{enumerate}

\paragraph{Construction.}
We propose a comprehensive framework for developing both the taxonomy and the \textsc{IPV-Txt} benchmark.
During the development process, the taxonomy structure and text prompts evolve together, informing and refining each other.
Specifically, our methodology consists of the following steps:

\textbf{1) Initialization.}
We establish the initial taxonomy structure by reviewing relevant literature~\cite{bansal2024videophy,phygenbench,meng2024phybench,zheng2024contphy} and leveraging large language models (LLMs), such as GPT-4o, for additional insights.
Beyond the widely studied domain of physical laws, we expand our taxonomy to include broader categories, incorporating everyday scenarios and intriguing effects that can be applied in downstream applications.

\textbf{2) Iterative Refinement.}
We refine the taxonomy and prompts through an iterative process.
At each iteration, we collect a set of text prompts to expand the existing taxonomy structure.
If certain text prompts do not fit within the structure, this suggests the need for adjustments to the taxonomy, ensuring that both the taxonomy and prompts evolve together.
New prompts drive the taxonomy toward greater comprehensiveness, while the updated taxonomy encourages human to explore more intriguing examples.

\textit{LLM Refinement.}
We carefully design prompts to guide various LLMs in generating a broader range of example scenarios.
We encourage the models to generate original and creative examples rather than merely substituting nouns in existing examples.
We employ three LLMs—namely GPT-4o, Claude-3.5, and Gemini-1.5—to maximize diversity.

\begin{figure}[t!]
    \centering
    \includegraphics[width=0.95\linewidth]{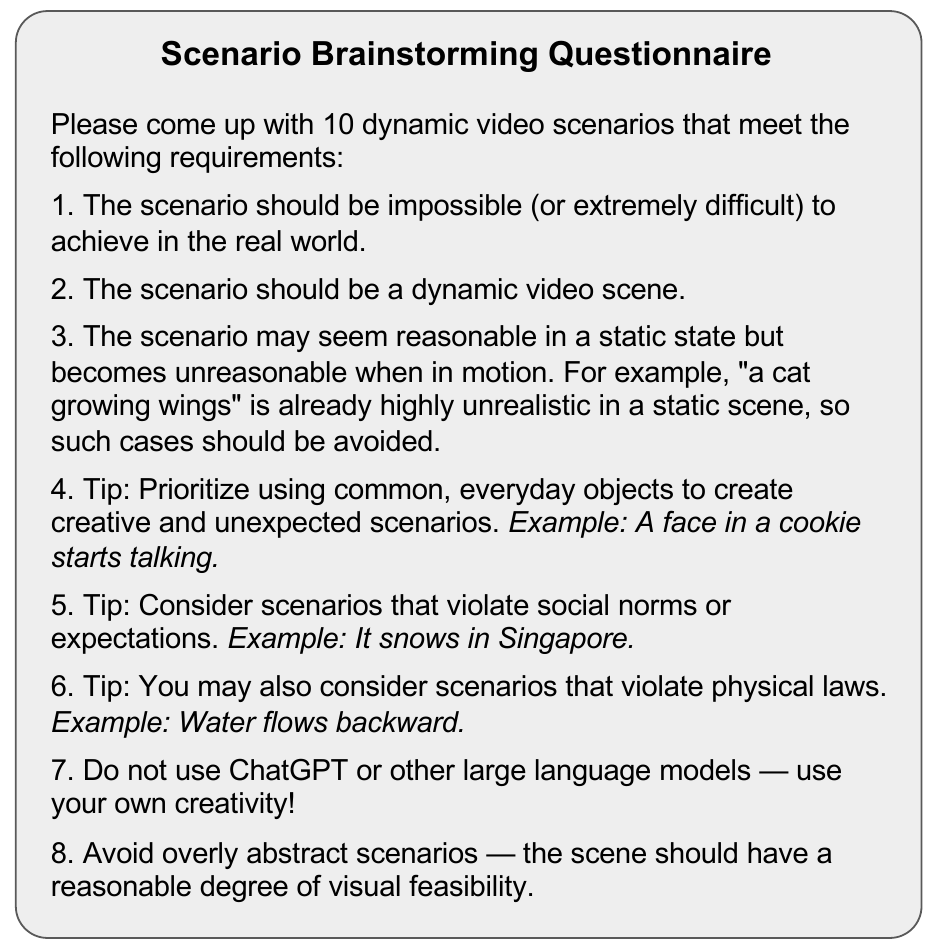}
    \vspace{-12pt}
    \caption{
    Questionnaire used for collecting impossible text prompts for \textsc{IPV-Txt}.
    }
    \vspace{-20pt}
    \label{fig:question_crowd}
\end{figure}

\textit{Crowdsourcing Refinement.}
To further improve the taxonomy, we aim to collect more ``outlier'' samples to challenge and refine the current structure until it accommodates the majority of samples.
To this end, we employ crowdsourcing to collect data samples from a diverse set of contributors.
As shown in Fig.~\ref{fig:question_crowd}, we employ a questionnaire to collect data from 15 participants from diverse academic and professional backgrounds, including computer science, economics, arts, and education.
The participants have acknowledged that the collected data will be used for research purposes.
On average, participants required 31 minutes to complete the questionnaire.
This time commitment demonstrates the complexity and value of the collected text prompts.
Finally, we have collected 150 impossible text prompts, which contribute to refining the taxonomy.
It is essential to explicitly specify ``LLMs are not allowed'' in the questionnaire to guarantee the authenticity of collected data samples.

\textbf{3) Quality Control.}
We conduct a thorough review of the prompts across multiple dimensions.
\textit{Clarity.} The text prompts should be expressed in a clear and comprehensible manner. Complicated or confusing prompts will be manually revised or removed.
\textit{Accessibility.} The described scene should be accessible in daily life. We exclude some obscure or esoteric samples.
\textit{Relevance.} The text prompt should be relevant to its belong category (and sub-category) in the taxonomy. We also merge some similar or duplicated samples at this step.
\textit{Visualizability.} The described scene should be able to be displayed by a video. Some non-visual scenes, such as sound-related descriptions, will be excluded.

\textbf{4) Linguistic Enhancement.}
We further refine the linguistic clarity and expressiveness of the text prompts to enhance their suitability for video generation models.
Specifically, we reference popular prompt rewriting strategies~\cite{kong2024hunyuanvideo,yang2024cogvideox} and design a prompt to instruct GPT-4o to enhance the text prompts.
Compared to the original descriptions, the rewritten prompts feature relatively longer sequences, greater detail, and a more structured format.

\begin{figure}[t!]
    \centering
    \includegraphics[width=0.9\linewidth]{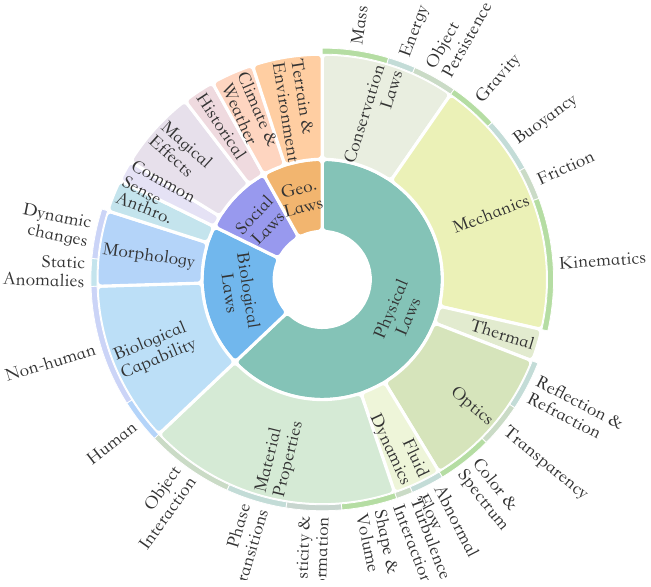}
    \vspace{-12pt}
    \caption{
    Distribution of the Prompt Suite Across the Taxonomy.
    }
    \vspace{-15pt}
    \label{fig:prompt_suite}
\end{figure}

This methodology yields a robust and comprehensive taxonomy, and a prompt suite, which can be used for assessing T2V models' capability on generating impossible videos, providing a valuable tool for advancing this domain. 
The distribution of the prompt suite is shown in Fig.~\ref{fig:prompt_suite}.

\subsection{\textsc{IPV-Vid}}
We construct \textsc{IPV-Vid}, a novel video benchmark designed to assess the capabilities of popular VideoLLMs in reasoning about temporal dynamics and world knowledge using impossible videos.
\textsc{IPV-Vid} is structured according to the established taxonomy.
Given the unique nature of video data, \textsc{IPV-Vid} places particular emphasis on temporal anomalies, such as motion and stage changes, which are challenging to identify from individual frames alone.

\subsubsection{Video Collection.}
\textbf{T2V Generation.}
We prompt 10 state-of-the-art T2V models to generate a comprehensive set of videos, including open-sourced models (Open-Sora~\cite{open_sora}, HunyuanVideo~\cite{kong2024hunyuanvideo}, CogVidX~\cite{yang2024cogvideox}, Mochi 1~\cite{genmo2025mochi1}, LTX~\cite{genmo2025mochi1}, and Pyramid-Flow~\cite{pyramid_flow}) and closed-sourced models (Sora~\cite{sora}, Kling~\cite{klingai}, Luma~\cite{luma}, and Hailuo~\cite{hailuoai}).
The \textsc{IPV-Txt} prompt suite is utilized as text prompts in text-to-video generation.
Using 260 high-quality text prompts from the \textsc{IPV-Txt} suite, we generate a total of 2,600 synthetic videos.

\textbf{Web Video Collection.}
To enhance the scale and diversity of the benchmark, we supplement our dataset with videos collected from the Internet.
The primary sources include community websites for commercial video generation models, such as Sora and Hailuo, among others.
Additionally, we gather videos from Twitter (X) shared by users, explicitly mentioning that they were generated by specific AI models.
During the manual collection process, we adhere to an implicit criterion: videos must exhibit phenomena that are impossible in the real world.
From these sources, we collect a total of 155 videos.

\textbf{Real-world Videos.}
To mitigate evaluation bias and ensure a balanced distribution, we incorporate real-world videos into the benchmark.
These videos are used alongside synthetic videos to evaluate the performance of VideoLLMs.
We utilize OpenVid~\cite{nan2024openvid} as the foundational dataset for real-world videos.
Using AI-generated videos as queries, we leverage the CLIP~\cite{clip} model to retrieve content-consistent videos from OpenVid.
During retrieval, we apply filtering criteria based on video length, aspect ratio, and aesthetic score to exclude unsuitable videos.
Additionally, we exclude videos with cartoon-style visuals, conspicuous logos, subtitles, and similar features.
In total, we collect 650 real-world videos for inclusion in the benchmark.

\begin{figure}[t!]
    \centering
    \includegraphics[width=0.88\linewidth]{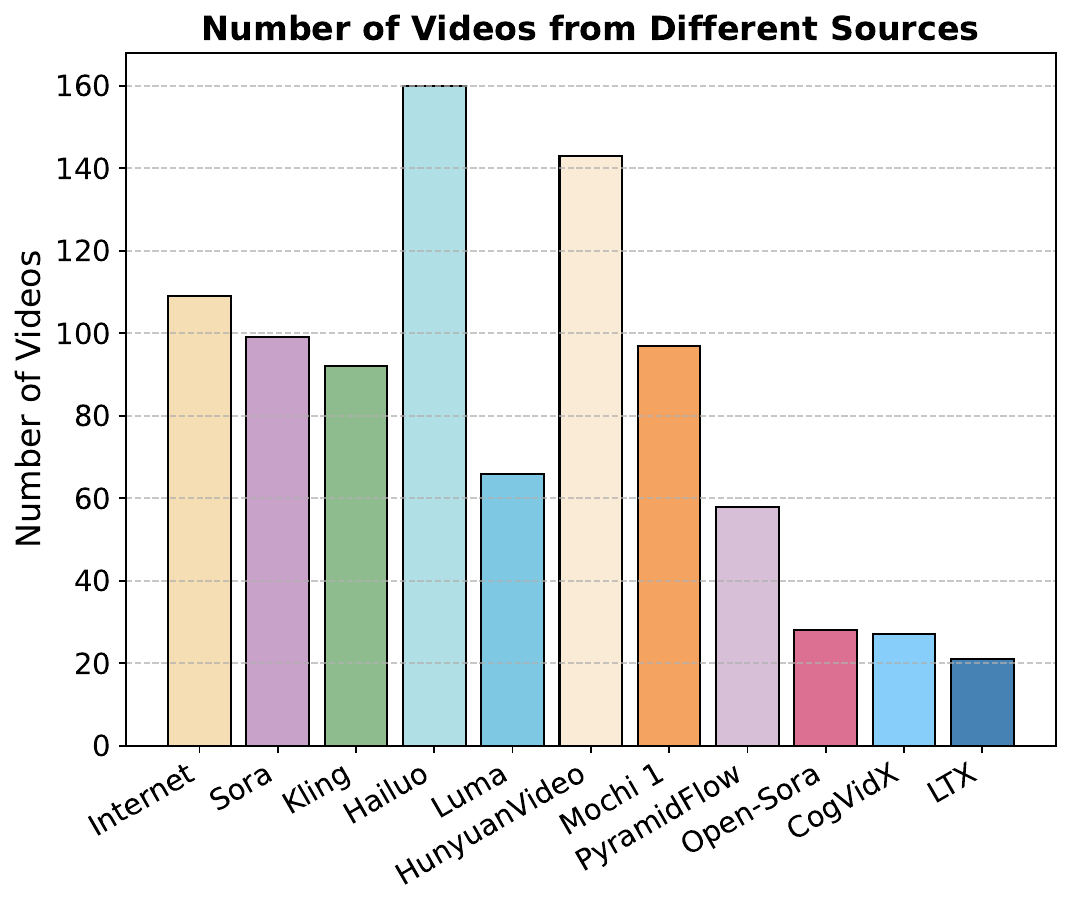}
    \vspace{-12pt}
    \caption{
    Sources of Impossible Videos.
    }
    \vspace{-15pt}
    \label{fig:ipv_vid_dist}
\end{figure}

\subsubsection{Human Annotation.}

\noindent\textbf{Step 1: Video Filtering.}
Our objective is to curate a collection of high-quality videos that depict impossible or counterfactual scenes.
To ensure data quality, we develop a custom annotation tool and perform human annotation on the collected videos.
A detailed description of the annotation tool is provided in the Appendix~\ref{supp:video_anno_tool}.
The main criteria of \textsc{IPV-Vid} contains two aspects:

\textit{1) Visual Quality}.
Videos of low visual quality are excluded, including but not limited to spatial blurring, temporal flickering, poor aesthetic quality, insufficient motion, and inconsistent style.

\textit{2) Semantic Plausibility}.
In line with the benchmark's objective, we retain videos with semantically counterfactual content—scenes that are impossible in the real world.

This ensures that the selected videos exhibit high visual quality while maintaining low semantic plausibility, thereby requiring models to reason based on semantic content rather than low-level visual features.
Fig.~\ref{fig:ipv_vid_dist} illustrates the distribution of retained impossible videos following the filtering process.

\noindent\textbf{Step 2: Detailed Annotation.}
For videos satisfying the aforementioned criteria, we perform detailed annotation with the following labels:

\textit{1) Spatial or Temporal Anomaly}.
This field requires annotators to determine whether the impossibility can be identified through spatial semantic information or necessitates temporal reasoning;

\textit{2) Taxonomy Category}.
Annotators assign a category label based on the IPV taxonomy;

\textit{3) Explanation}.
Annotators provide a brief textual description of the specific impossible phenomenon depicted in the video.

\subsubsection{Task Design.}

\noindent\textbf{Judgment Task} requires models to classify the input video as either synthetic or real by answering the question, ``Is the provided video generated by AI?"
To minimize the influence of visual elements and style, ensuring models focus on semantic content, we use synthetic videos without watermarks and exclude real-world videos with cartoon-style visuals, conspicuous logos, subtitles, or similar features.
To ensure a balanced evaluation, the dataset maintains a $1:1$ ratio of synthetic to real-world videos.
This task is framed as a binary classification problem and evaluated using average Accuracy and F1-score.
Additionally, we report the `yes rate' in Appendix~\ref{supp:aigc_judge} to facilitate model diagnosis.

\begin{table*}[ht]
\centering
\caption{
\textbf{Evaluation Results of \textsc{IPV-Txt} Across Dimensions.}
This table compares the performance of state-of-the-art video generation models using the \textsc{IPV-Txt} benchmark as text prompts in the T2V setting. A higher score indicates better performance in a given dimension.
\textbf{Bold} denotes best, \underline{underline} denotes second.
}
\vspace{-8pt}
\resizebox{0.9\linewidth}{!}{
\begin{tabular}{lccccccccccc}
\toprule
\multirow{3}{*}{\textbf{Model}} & \multicolumn{2}{c}{\textbf{Physical}} & \multicolumn{2}{c}{\textbf{Biological}} & \multicolumn{2}{c}{\textbf{Social}} & \multicolumn{2}{c}{\textbf{Geographical}} & \multicolumn{3}{c}{\textbf{Overall}} \\ 
\cmidrule(lr){2-3} \cmidrule(lr){4-5} \cmidrule(lr){6-7} \cmidrule(lr){8-9} \cmidrule(lr){10-12}
      & Visual & Prompt & Visual & Prompt & Visual & Prompt & Visual & Prompt & Visual & Prompt & IPV \\
      & Quality & Following & Quality & Following & Quality & Following & Quality & Following & Quality & Following & Score \\
\midrule
\hline
\rowcolor{mygray}
\multicolumn{12}{l}{\emph{Open-source Models}} \\
\hline
LTX~\cite{hacohen2024ltx}  & 58.3 & 14.1 & 35.3 & 21.6 & 44.0 & 16.0 & 57.1 & 23.8 & 52.3 & 16.5 & 10.0 \\
Open-Sora~\cite{open_sora}  & 63.8 & 22.1 & 25.5 & 29.4 & 20.0 & 32.0 & 57.1 & 47.6 & 51.5 & 26.5 & 15.8 \\ 
Pyramid-Flow~\cite{pyramid_flow} & 88.3 & 15.3 & 92.2 & 21.6 & 72.0 & 28.0 & \textbf{100.0} & 52.4 & 88.5 & 20.8 & 19.6 \\
CogVidX-1.5~\cite{yang2024cogvideox}  & 38.7 & 29.4 & 35.3 & \underline{52.9} & 40.0 & \underline{56.0} & 81.0 & \underline{61.9} & 41.5 & \underline{39.2} & 16.9 \\
Mochi 1~\cite{genmo2025mochi1} & 68.7 & \textbf{44.2} & 64.7 & \textbf{56.6} & 80.0 & \textbf{60.0} & 71.4 & \textbf{76.2} & 69.2 & \textbf{50.8} & \textbf{37.3} \\
HunyuanVid~\cite{kong2024hunyuanvideo} & 95.1 & 23.9 & 88.2 & 35.3 & 88.0 & 40.0 & 90.5 & 42.9 & 92.7 & 29.2 & 26.2 \\ 
\hline
\rowcolor{mygray}
\multicolumn{12}{l}{\emph{Proprietary Models}} \\
\hline
\bottomrule

Luma~\cite{luma} & 88.3 & 11.7 & 90.2 & 19.6 & 82.6 & 17.4 & 85.7 & 52.4 & 88.0 & 17.1 & 14.3 \\ 
Sora~\cite{sora} & \textbf{98.8} & 15.3 &\textbf{98.0} & 43.1 & \textbf{100.0} & 30.4 & \underline{95.2} & \underline{61.9} & \textbf{98.4} & 26.0 & 25.2 \\ 
Kling~\cite{klingai} & \textbf{98.8} & 21.5 & \underline{94.1} & 33.3 & \underline{95.7} & 43.5 & 81.0 & 42.9 & \underline{96.1} & 27.5 & 26.7 \\ 
Hailuo~\cite{hailuoai} & \underline{96.3} & \underline{30.1} & \underline{94.1} & 45.1 & \textbf{100.0} & 52.0 & 90.5 & \underline{61.9} & 95.8 & 37.7 & \underline{36.2} \\
\bottomrule

\end{tabular}
}
\vspace{-10pt}
\label{tab:performance_metrics}
\end{table*}

\noindent\textbf{Multi Choice Task (MCQA)} task requires models to identify the description that best captures the impossible phenomenon depicted in the video.
The question is formulated as follows: `Select the best answer to the following multiple-choice question based on the video'.
To create effective distractors that challenge the model, we carefully design an instructional prompt and leverage the GPT-4o model for distractor generation.

The instructional prompt is designed with the following considerations:
1) Ensure all options, including both correct answer and distractors, are similar in length, style, detail degree, and complexity;
2) Ensure distractors also present specific impossible phenomenon;
3) Ensure the impossible phenomenon in distractor shall involve visual elements shown in the given video frame, to avoid the model solve the problem with simple visual element grounding.
By incorporating detailed annotations and visual content as references, we mitigate hallucination in GPT-4o, ensuring high-quality distractor generation.
A qualitative example of distractors is provided in Fig.~\ref{fig:mcqa_example}.
The instructional prompt is included in Appendix~\ref{supp:mcqa_prompt}.
The MCQA task is treated as a multi-class classification problem and evaluated using mean Accuracy.

\noindent\textbf{Open-ended QA Task (OpenQA).}
We introduce an open-ended impossible explanation task, which requires models to \textit{independently} and \textit{correctly} identify the impossible phenomenon depicted in the video without any hints.
The task is framed with the following question: ``Based on your observation of the video, what content or event makes the video impossible or unusual in the real world?''.
Compared to the MCQA task, the open-ended explanation task is more challenging, as models must generate responses without reference to candidate options.
This task more accurately assesses whether models can genuinely perceive and articulate detailed anomalies, rather than relying on guesswork.

For evaluation, we employ an LLM as an evaluator to score model responses by comparing them to the annotated text explanations in the benchmark.
Empirically, we observe that directly instructing the LLM to assign scores results in instability.
To address this, we propose a \textit{justification-then-score} approach.
In this approach, the evaluator first provides a justification by identifying key matches or mismatches between the model's prediction and the ground truth.
Based on this justification, the evaluator assigns a semantic alignment score on a scale from 0 to 1, where 1.0 indicates perfect alignment, 0.8-0.9 indicates good alignment, 0.5-0.7 indicates partial alignment, 0.1-0.4 indicates weak alignment, and 0.0 indicates no alignment.
The justification step is critical to ensure fair and stable score assignment.

By default, we employ GPT-4o as the primary evaluator.
To avoid self-evaluation bias, as GPT-4o is also evaluated as a VideoLLM, we additionally utilize Claude-3.5-Sonnet as an evaluator.

\noindent\textbf{Quality Assessment.}
Since the task of MCQA and OpenQA is constructed or evaluated with the help of GPT-4o, which risks hallucination, we conduct a human assessment.
Specifically, for each task, we randomly select a subset of 100 samples and ask human to answer the questions, imitating the video understanding models.
After that, the accuracy or score serve as a golden reference in models' evaluation.

\vspace{-5pt}
\section{Evaluate Impossible Video Generation}
\label{sec:eval_vid_gen}
\vspace{-2pt}

\noindent \textbf{Setup.}
We evaluate mainstream video generation models, including open-source (Open-Sora 1.2~\citep{open_sora}, HunyuanVideo~\citep{kong2024hunyuanvideo}, CogVidX~\citep{yang2024cogvideox}, Mochi 1~\citep{genmo2025mochi1}, LTX~\citep{hacohen2024ltx}, Pyramid-Flow~\citep{pyramid_flow}) and closed-source models (Sora~\citep{sora}, Kling 1.5~\citep{klingai}, Luma~\citep{luma}, and Hailuo~\citep{hailuoai}).

\noindent \textbf{IPV-Score Metric.}
To evaluate a model, we first generate a set of videos using the \textsc{IPV-Txt} prompt suite.
Human annotators are then employed to label two aspects of each video:
1) \textit{Visual quality}, assessing whether the video meets high-quality standards,
and 2) \textit{Impossible prompt following}, determining whether the video accurately depicts the impossible event described in the text prompt.
Annotators provide binary labels for each dimension.
Using these labels, we introduce the IPV-Score, a novel metric to evaluate a model’s ability to generate high-quality videos that faithfully depict impossible events.
The IPV-Score is calculated as the percentage of videos that are both high-quality and faithful to the prompt within the entire generated set.

\noindent \textbf{Human- and Auto-eval.}
Tab.~\ref{tab:performance_metrics} presents the performance of the evaluated models.
Additionally, we propose an automated evaluation method to assess a model’s capability in generating impossible videos.
A detailed analysis is provided in the Appendix~\ref{supp:auto_eval}.

\vspace{-8pt}
\subsection{Results and Analysis}
\vspace{-5pt}

\noindent\textbf{Can today's video generation models effectively follow prompts to create impossible video content?}
As shown in Tab.~\ref{tab:performance_metrics}, the top-performing model, Mochi 1, generates high-quality impossible videos in only $37.3\%$ of cases.
Other models perform even worse, suggesting that current video generation models remain far from achieving satisfactory performance in generating high-quality impossible videos.

\noindent\textbf{Unbalanced capability between visual quality and impossible prompt following.}
For instance, Luma demonstrates remarkable visual quality, surpassing most open-source models, but its prompt-following score is significantly lower.
In contrast, some open-source models, such as Mochi 1, exhibit superior prompt-following capabilities, even outperforming many proprietary models.
An ideal model should excel in both dimensions, achieving a balance that is quantified by our IPV-Score metric.

\begin{figure}[t!]
    \centering
    \includegraphics[width=0.95\linewidth]{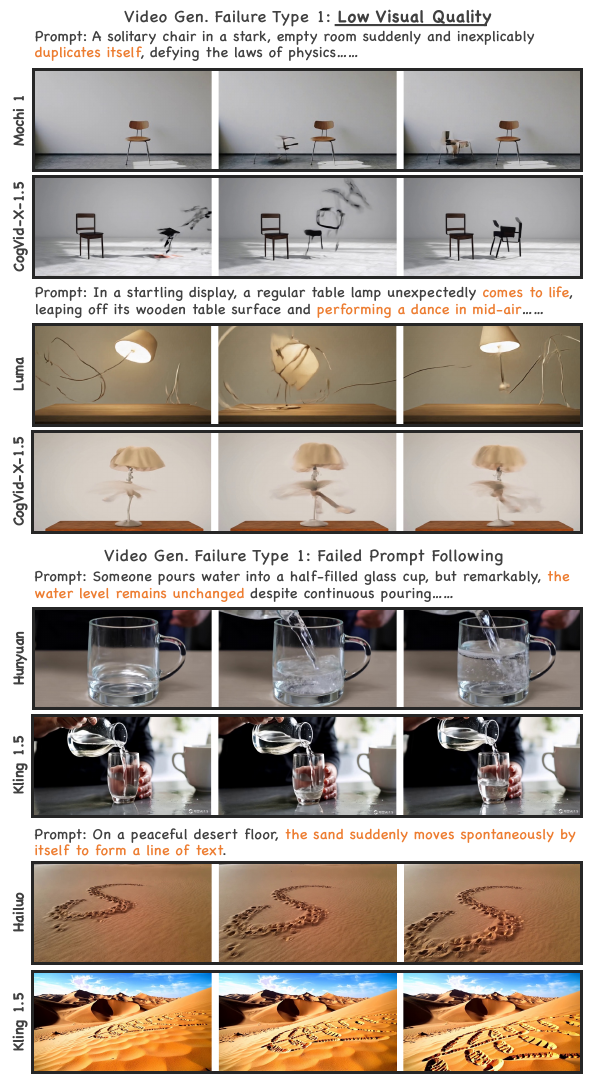}
    \vspace{-12pt}
    \caption{
    Failure case of impossible video generation.
    }
    \vspace{-15pt}
    \label{fig:vid_gen_fail}
\end{figure}

\noindent\textbf{What limits the the model on impossible video generation?}
Beyond basic prompt comprehension and following, we identify two unique challenges posed by impossible text prompts, as illustrated in Fig.~\ref{fig:vid_gen_fail}.
\textit{First, impossible prompts may trigger low visual quality.}
While the model attempts to follow the prompt, the unusual nature of impossible prompts often results in visual artifacts or generation failures.
This is likely because impossible prompts represent out-of-distribution data for the model.
\textit{Second, an overemphasis on adhering to physical laws may constrain the model's creative freedom.}
In many failure cases, videos accurately capture the semantic elements of the prompt but fail to depict the critical impossible phenomenon.
Instead, they depict normal scenes that conform to real-world rules.
Future video generation models may consider the above factors to achieve a better model.
Additional examples are included in the Appendix~\ref{supp:cast_study_gen}.

\section{Evaluate Impossible Video Understanding }
\label{sec:eval_vid_llm}

\begin{table*}[ht]
\centering
\caption{
\textbf{Evaluation Results for Impossible Video Understanding.}
This table compares the performance of sota VideoLLMs using the \textit{IPV-Vid} benchmark. A higher score indicates better performance in a given dimension.
\textbf{Bold} denotes best, \underline{underline} denotes second, Open(C) denotes using Claude as the evaluator.
}
\vspace{-8pt}
\resizebox{0.9\linewidth}{!}{
\begin{tabular}{l|cc|ccccccccccc}

\toprule

\multirow{2}{*}{\textbf{Model}} & \multicolumn{2}{c|}{\textbf{Judgement}} & \multicolumn{2}{c}{\textbf{Physical}} & \multicolumn{2}{c}{\textbf{Biological}} & \multicolumn{2}{c}{\textbf{Social}} & \multicolumn{2}{c}{\textbf{Geographical}} & \multicolumn{3}{c}{\textbf{Overall}} \\ 

\cmidrule(lr){2-3} \cmidrule(lr){4-5} \cmidrule(lr){6-7} \cmidrule(lr){8-9} \cmidrule(lr){10-11} \cmidrule(lr){12-14}
      & Acc. & F1 & MC & Open & MC & Open & MC & Open & MC & Open & MC & Open & Open(C) \\
\midrule
\hline
\rowcolor{mygray}
\multicolumn{14}{l}{\emph{Open-source Models}} \\
\hline

Random & 50.0 & 50.0 & 20.0 & - & 20.0 & - & 20.0 & - & 20.0 & - & 20.0 & - & - \\
Human & - & - & - & - & - & - & - & - & - & - & \textcolor{gray}{94.0} & \textcolor{gray}{82.7} & - \\

Video-LLaVA~\cite{video_llava}      & 72.7 & \textbf{72.9} & 23.0 & 13.3 & 34.6 & 29.7 & 31.8 & 16.9 & 24.1 & 25.9 & 26.8 & 14.2 & 18.7  \\
Oryx~\cite{liu2024oryx}             & 58.6 & 44.6 & 52.0 & 16.7 & 68.8 & 33.9 & 70.3 & 32.6 & 83.9 & 35.2 & 60.4 & 18.7 & 22.7 \\
Intern-VL-2.5~\cite{intern_vl25}    & 56.5 & 69.7 & 61.9 & 34.9 & 64.2 & 59.6 & 65.5 & 49.6 & 77.0 & 48.4 & 62.4 & 33.0 & 32.5 \\
NVILA~\cite{liu2024nvila}           & 72.6 & 64.0 & 60.2 & 26.4 & 63.3 & 52.3 & 68.9 & 36.6 & 69.0 & 42.9 & 62.6 & 26.8 & 30.6 \\
LongVU~\cite{shen2024longvu}        & 70.3 & 68.0 & 69.3 & 21.8 & 79.6 & 38.3 & 77.0 & 35.6 & 77.0 & 35.6 & 73.3 & 21.9 & 25.4 \\
Qwen2-VL~\cite{qwen2_vl}            & \textbf{76.2} & \underline{71.1} & 69.1 & 32.8 & 75.8 & 56.3 & 75.0 & 48.3 & 75.9 & 53.0 & 71.4 & 31.7 & 33.7 \\
LLaVA-Next~\cite{llava_next}        & \underline{73.2} & 70.4 & \textbf{82.8} & 37.7 & \textbf{92.9} & 57.2 & \textbf{90.5} & 51.3 & \underline{90.8} & 51.4 & \textbf{86.4} & 34.4 & \underline{38.6} \\

\hline
\rowcolor{mygray}
\multicolumn{14}{l}{\emph{Proprietary Models}} \\
\hline
\bottomrule
Gemini-1.5-Flash~\cite{gemini15} & 73.1 & 64.0 & \underline{80.5} & \underline{48.5} & \underline{90.8} & \underline{66.4} & \underline{89.2} & \underline{59.3} & \textbf{93.1} & \underline{65.3} & \underline{84.4} & \underline{42.5} & 36.8 \\
GPT-4o~\cite{gpt4o} & - & - & 76.7 & \textbf{58.2} & 83.8 & \textbf{75.5} & 84.5 & \textbf{64.9} & 92.0 & \textbf{71.1} & 79.7 & \textbf{49.1} & \textbf{45.1} \\

\bottomrule

\end{tabular}
}
\vspace{-10pt}
\label{tab:vid_llm_eval_main}
\end{table*}

We evaluate a range of popular Video-LLMs, encompassing both open-source and proprietary models.
The three tasks are designed to form a hierarchy, increasing in difficulty to comprehensively evaluate model capabilities.
Tab.~\ref{tab:vid_llm_eval_main} presents the evaluation result of impossible video understanding across models.

\begin{figure}[t!]
    \centering
    \includegraphics[width=0.95\linewidth]{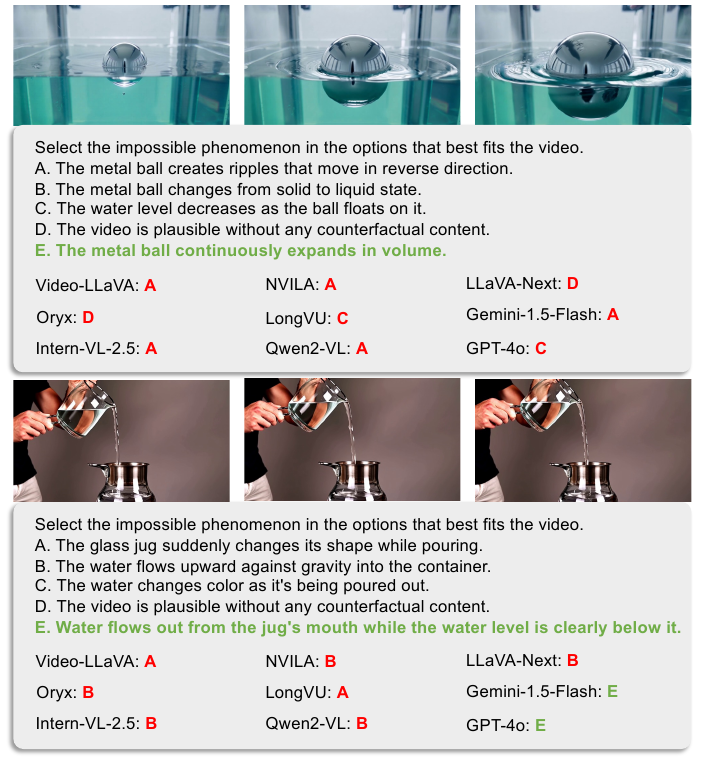}
    \vspace{-12pt}
    \caption{
    Example of the MCQA task.
    We highlight the correct option in \textbf{\textcolor{red}{red}} and the incorrect option in \textbf{\textcolor[rgb]{0.0, 0.5, 0.0}{green}}.
    }
    \vspace{-15pt}
    \label{fig:mcqa_example}
\end{figure}

\vspace{-8pt}
\subsection{Results}
\vspace{-5pt}

\noindent\textbf{Judgment Task.}
Most models achieve comparable Accuracy and F1-scores on this task.
Qwen2-VL achieves the highest accuracy at $76.2\%$, even outperforming Gemini by $3.1$ percentage points.
Empirically, we observe that some models exhibit bias, predominantly answering `Yes' (or `No') for the majority of videos.
To account for this bias, we also report F1-scores, with Video-LLaVA achieving the highest performance.
Detailed results and additional analysis are provided in the Appendix~\ref{supp:aigc_judge}.
GPT-4o reject answering ``Yes'' or ``No'' for this task.
To ensure fair comparison, we do not further tune the text prompt for it.

\begin{table}[ht]
\centering
\caption{
Video understanding evaluation results on two categories of videos: 1) videos that can be understood through \textbf{spatial} scene understanding and world knowledge, and 2) videos that require \textbf{temporal} reasoning for comprehension. 
}
\vspace{-8pt}
\resizebox{0.85\linewidth}{!}{
\begin{tabular}{lcccc}

\toprule

\multirow{2}{*}{\textbf{Model}} & \multicolumn{2}{c}{\textbf{Spatial}} & \multicolumn{2}{c}{\textbf{Temporal}} \\ 

\cmidrule(lr){2-3} \cmidrule(lr){4-5}
      & MC & Open & MC & Open \\
\midrule
\hline
\rowcolor{mygray}
\multicolumn{5}{l}{\emph{Open-source Models}} \\
\hline

Random & 20.0 & - & 20.0 & - \\
Video-LLaVA~\cite{video_llava} & 34.6 & 30.3 & 23.8 & 13.7 \\
Oryx~\cite{liu2024oryx} & 83.2 & 41.0 & 51.4 & 17.5 \\
Intern-VL-2.5~\cite{intern_vl25} & 72.4 & 56.6 & 60.6 & 37.2 \\
NVILA~\cite{liu2024nvila} & 75.2 & 51.6 & 57.6 & 28.0 \\
LongVU~\cite{shen2024longvu} & 85.7 & 42.6 & 68.5 & 22.7 \\
Qwen2-VL~\cite{qwen2_vl} & 79.7 & 57.6 & 68.2 & 34.4 \\
LLaVA-Next~\cite{llava_next} & 95.5 & 55.8 & 82.7 & 39.8 \\

\hline
\rowcolor{mygray}
\multicolumn{5}{l}{\emph{Proprietary Models}} \\
\hline
\bottomrule

Gemini-1.5-Flash~\cite{gemini15} & 93.0 & 67.5 & 81.1 & 50.0 \\
GPT-4o~\cite{gpt4o} & 90.6 & 75.2 & 75.6 & 59.1 \\

\bottomrule

\end{tabular}
}
\vspace{-17pt}
\label{tab:vid_llm_eval_st}
\end{table}

\noindent\textbf{Multi Choice Task.}
Model performance on this task exhibits significant variation.
The top-performing model, LLaVA-Next, achieves an accuracy of $86.4\%$, surpassing both GPT-4o and Gemini.
In contrast, Video-LLaVA achieves only $26.8\%$ accuracy, which is close to the random baseline.
Most open-source models exhibit substantial room for improvement.
Fig.~\ref{fig:mcqa_example} presents an example of the MCQA task, where most models fail to select the correct option.

\begin{figure}[t!]
    \centering
    \includegraphics[width=0.99\linewidth]{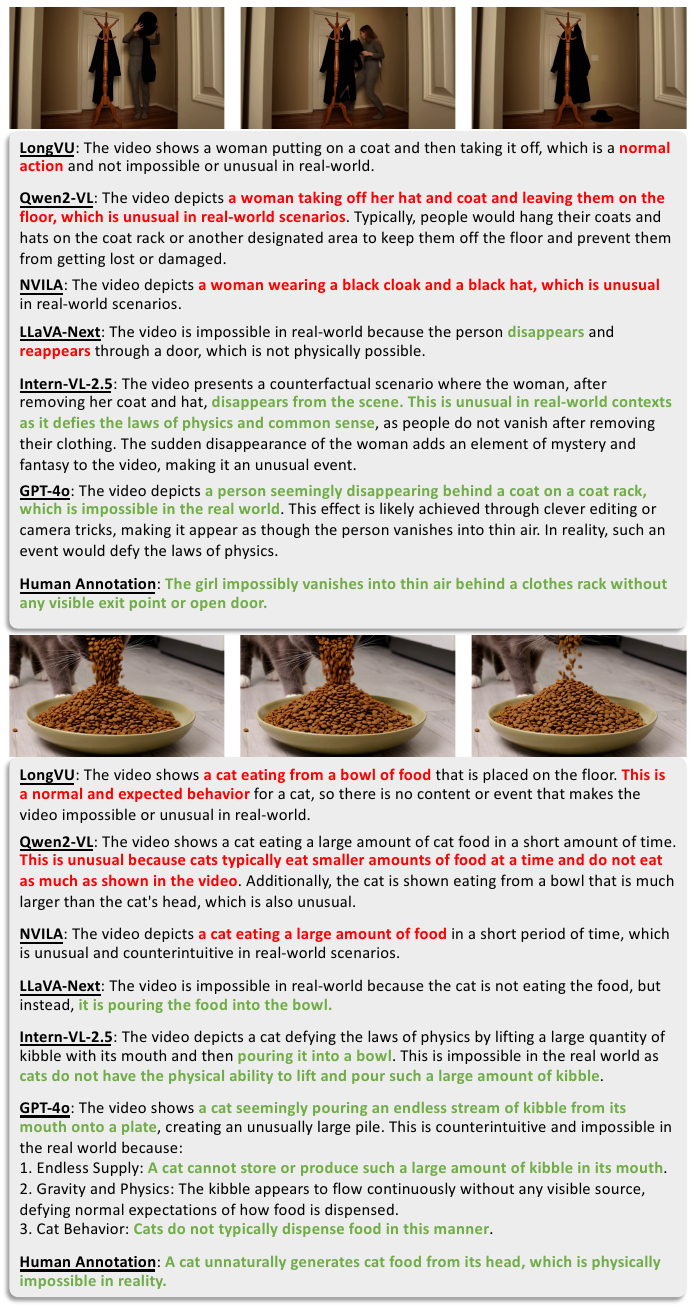}
    \vspace{-12pt}
    \caption{
    Example of the OpenQA task.
    We ask state-of-the-art video understanding models to analyze whether the video is impossible or not.
    We highlight the correct analysis in \textbf{\textcolor{red}{red}} and the incorrect analysis in \textbf{\textcolor[rgb]{0.0, 0.5, 0.0}{green}}.
    }
    \vspace{-15pt}
    \label{fig:openqa_example}
\end{figure}

\noindent\textbf{Open-ended QA Task.}
Among the three tasks, the OpenQA task is the most conceptually straightforward yet practically challenging.
The last two column of Tab.~\ref{tab:vid_llm_eval_main} indicate that most current models struggle to independently identify counterfactual phenomena in videos.
Among open-source models, LLaVA-Next achieves superior scores with both GPT-4o and Claude-3.5 evaluators, aligning with its strong performance on the MCQA task.
GPT-4o attains the highest performance with both evaluators, underscoring its robust visual understanding and reasoning capabilities.
The overall scores suggest that most models struggle to comprehend impossible videos, highlighting a promising direction for future research.
Fig.~\ref{fig:openqa_example} illustrates an example of the OpenQA task.

Comparing the results from the two evaluators, we observe that GPT-4o tends to be stricter, assigning lower scores than Claude-3.5 for most models.
However, when evaluating GPT-4o's performance on video understanding, the GPT-4o evaluator assigns higher scores than Claude-3.5, suggesting potential self-evaluation bias.
Despite this, the discrepancy between evaluators does not alter the conclusion that GPT-4o is the top-performing model for impossible video understanding.

\subsection{Analysis}

\noindent\textbf{Are today's video understanding models good enough for understanding impossible videos?}
The results from the MCQA and OpenQA tasks collectively provide insights into this question.
Overall, proprietary models show promising potential, outperforming most open-source models on this task.
However, their ability to independently identify impossible videos remains suboptimal, as evidenced by the OpenQA scores.
Most open-source models perform poorly on this task, indicating significant room for improvement.

\noindent\textbf{Unbalanced model capabilities across domains and tasks.}
Tab.~\ref{tab:vid_llm_eval_main} reveals that many models exhibit unbalanced capabilities across domains and tasks.
Across domains, ``\textit{Physical}'' emerges as the most challenging, with most models achieving the lowest scores in this category.
In contrast, the remaining 3 domains exhibit relatively balanced performance across models.
We hypothesize that ``\textit{Physical}'' contains more challenging samples that necessitates temporal dynamic reasoning.
Across tasks, performance on the MCQA and OpenQA tasks is closely correlated, whereas the Judgment task appears to be distinct.
For instance, Video-LLaVA excels in the Judgment task but underperforms significantly in the other tasks.
Discrepancies also exist between MCQA and OpenQA performance.
LLaVA-Next shows strong performance on MCQA but underperforms on OpenQA.
In contrast, GPT-4o achieves the highest performance on OpenQA but lags slightly behind on MCQA.

\noindent\textbf{What makes a good model for impossible video understanding?}
We identify two critical factors: temporal dynamic reasoning and world knowledge reasoning.
For instance, in the first example from Fig.~\ref{fig:teaser_vids}, the cookie appears plausible in individual frames.
However, only by analyzing and reasoning across frames can one detect the self-growing phenomenon—an impossible event.
Conversely, the last example in Fig.~\ref{fig:teaser_vids} ("snowing in Singapore") requires world knowledge for identification—specifically, the understanding that Singapore is a tropical country.
Due to space limit, additional case studies are provided in the Appendix~\ref{supp:cast_study_understanding}.

\noindent\textbf{Challenge and opportunity on temporal reasoning.}
Of the two key factors, world knowledge is primarily governed by the LLM, whereas temporal reasoning offers greater design flexibility but remains more challenging.
Tab.~\ref{tab:vid_llm_eval_st} provides a separate evaluation of spatial- and temporal-focused videos.
For all models, scores on temporal-focused videos are consistently lower than those on spatial-focused videos.
This clearly demonstrates that temporal dynamic reasoning poses significant challenges for most current models.
Video expert models with high frame rates (e.g., LongVU) do not exhibit a significant advantage.
Interestingly, the top-performing models (e.g., LLaVA-Next and GPT-4o) are all image-based.
It is worth noting that GPT-4o is evaluated using only 1 FPS.
This observation suggests that more sophisticated temporal modules, rather than simply expanding the context window, may be key to understanding and reasoning about impossible videos.

\vspace{-8pt}
\section{Conclusion}
\vspace{-5pt}
\label{sec:conclusion}

In this work, we introduce the concept of \textit{impossible videos} as a novel testbed to challenge and advance video understanding and generation models.
Unlike real-world videos, impossible videos present counterfactual and anti-reality scenarios, demanding models to go beyond mere memorization and retrieval, requiring deeper reasoning and generalization.
To facilitate research in this direction, we construct \textsc{IPV-Bench}, a comprehensive benchmark comprising a well-structured taxonomy, a diverse prompt suite (\textsc{IPV-Txt}), and a high-quality video dataset (\textsc{IPV-Vid}).
Through extensive evaluations, we demonstrate that current video models struggle with impossible videos, revealing significant gaps in their ability to reason about non-real-world scenarios. 
Our findings provide valuable insights into the limitations of existing models and highlight promising future research directions.

\nocite{langley00}

\clearpage
\bibliography{main}
\bibliographystyle{icml2025}

\newpage
\appendix

\onecolumn

\section{Additional Details of Benchmark Construction}

\subsection{Video Annotation Tool}
\label{supp:video_anno_tool}
\begin{figure*}[h]
    \centering
    \includegraphics[width=0.9\linewidth]{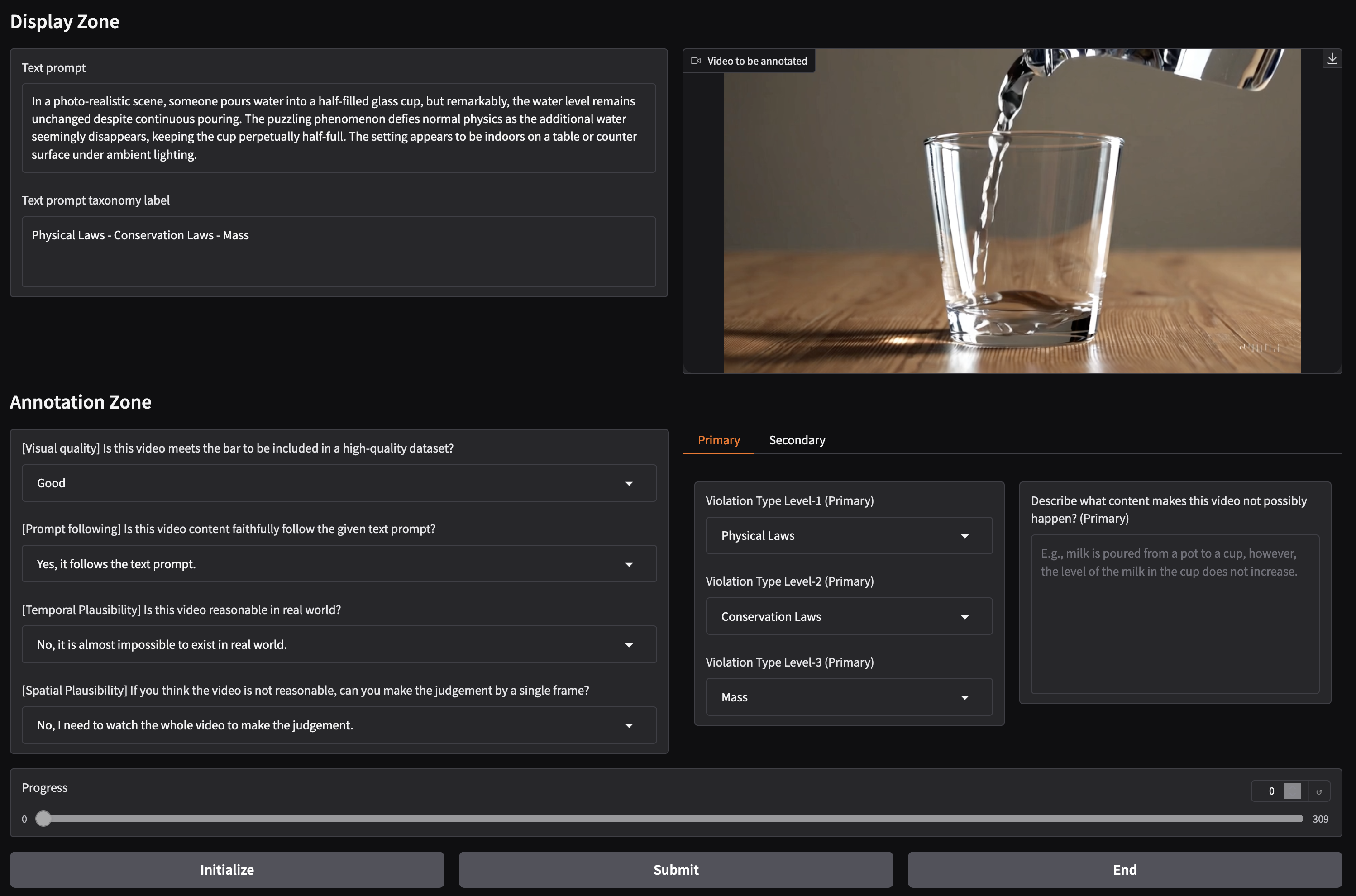}
    \vspace{-10pt}
    \caption{
    Screenshot of the annotation tool.
    }
    \label{fig:anno_tool}
\end{figure*}

Fig.~\ref{fig:anno_tool} illustrates a screenshot of our data annotation tool.
It has been divided into two zones:
Display Zone showing the essential information for annotation, including original text prompt, the taxonomy label of the prompt, and the video.
Annotation Zone provides several fields for annotations, including data for both \textsc{IPV-Vid} curation and labels for video generation models evaluation.
This all-in-one evaluation tool greatly eases the annotation efforts, supporting multiple ways for downstream usage.
Specifically, the questions for each filed is carefully designed to fit human customs.
For example, although we aim to annotate ``if this video is impossible'', we find it is more intuitive to answer ``if this video is reasonable'' in practice.

\section{Additional Results and Analysis}

\subsection{AI-Generated Video Judgment}
\label{supp:aigc_judge}
Tab.~\ref{tab:vid_llm_eval_binary} presents more detailed results on the task of AI-Generated video judgment, helping understanding unique properties of each model.
Video-LLaVA is a representative balanced model, with similar accuracy on both fake and real videos, and Yes Rate around $50\%$.
In contrast, Intern-VL, NVILA and Gemini are significantly biased.
Intern-VL is biased to answer `Yes' with a high Yes Rate $93.5\%$.
NVILA, and Gemini are biased toward the other direction, preferring answer `No' with lower Yes Rate.
The trends can also be reflected in the huge accuracy difference between fake and real videos.

\begin{table*}[ht]
\centering
\caption{
\textbf{Detailed Evaluation Results for AI-Generated Video Judgment Task.}
}
\resizebox{0.8\linewidth}{!}{
\begin{tabular}{lccccc}

\toprule

\textbf{Model} & \textbf{Overall Acc.} & \textbf{Fake Acc.} & \textbf{Real Acc.}  & \textbf{F1-Score} & \textbf{Yes Rate} \\ 
\hline

Video-LLaVA~\cite{video_bench} & 72.7 & 72.9 & 72.5 & 72.9 & 50.3 \\
Oryx~\cite{liu2024oryx} & 58.6 & 33.2 & 84.3 & 44.6 & 24.5 \\
Intern-VL-2.5~\cite{intern_vl25} & 56.5 & 99.7 & 12.8 & 69.7 & 93.5 \\
NVILA~\cite{genvidbench} & 72.6 & 48.4 & 97.1 & 64.0 & 25.8 \\
LongVU~\cite{shen2024longvu} & 70.3 & 62.7 & 78.0 & 68.0 & 42.5 \\
Qwen2-VL~\cite{qwen2_vl} & 76.2 & 58.3 & 94.3 & 71.1 & 32.1 \\
LLaVA-Next~\cite{llava_next} & 73.2 & 63.3 & 83.2 & 70.4 & 40.2 \\
Gemini-1.5-Flash~\cite{gemini15} & 73.1 & 47.6 & 98.8 & 64.0 & 24.6 \\

\bottomrule

\end{tabular}
}
\label{tab:vid_llm_eval_binary}
\end{table*}

\subsection{Evaluating Impossible Video Generation  -- An Automatic Strategy}
\label{supp:auto_eval}
In this section, we introduce an automatic evaluation strategy as a surrogate to human score presented in the main text. We first generate a collection of videos based on the IPV-Txt prompt suite for each model. Then, we compute the the following metrics to evaluate the visual quality and impossible prompt following capability, respectively.

\textit{1) Visual Quality.}
We use the popular video quality assessment suite, VBench~\citep{huang2024vbench}, to build a compound metric that measures overall video quality without considering the prompt. Specifically, we combine the six factors—Subject Consistency, Background Consistency, Motion Smoothness, Aesthetic Quality, Imaging Quality, and Dynamic Degree—from VBench to form our final metric. Similar to VBench, we calculate the weighted average of these six factors as the final evaluation score. We tailor the weights of these factors to better suit our video domain, such as reducing the weight of Aesthetic Quality since our impossible videos mostly follow a realistic style. The weights we use for each factor are: 2.0, 2.0, 0.2, 0.2, 2.0, 1.0.

\begin{figure}[t]
    \centering
    \includegraphics[width=0.66\linewidth]{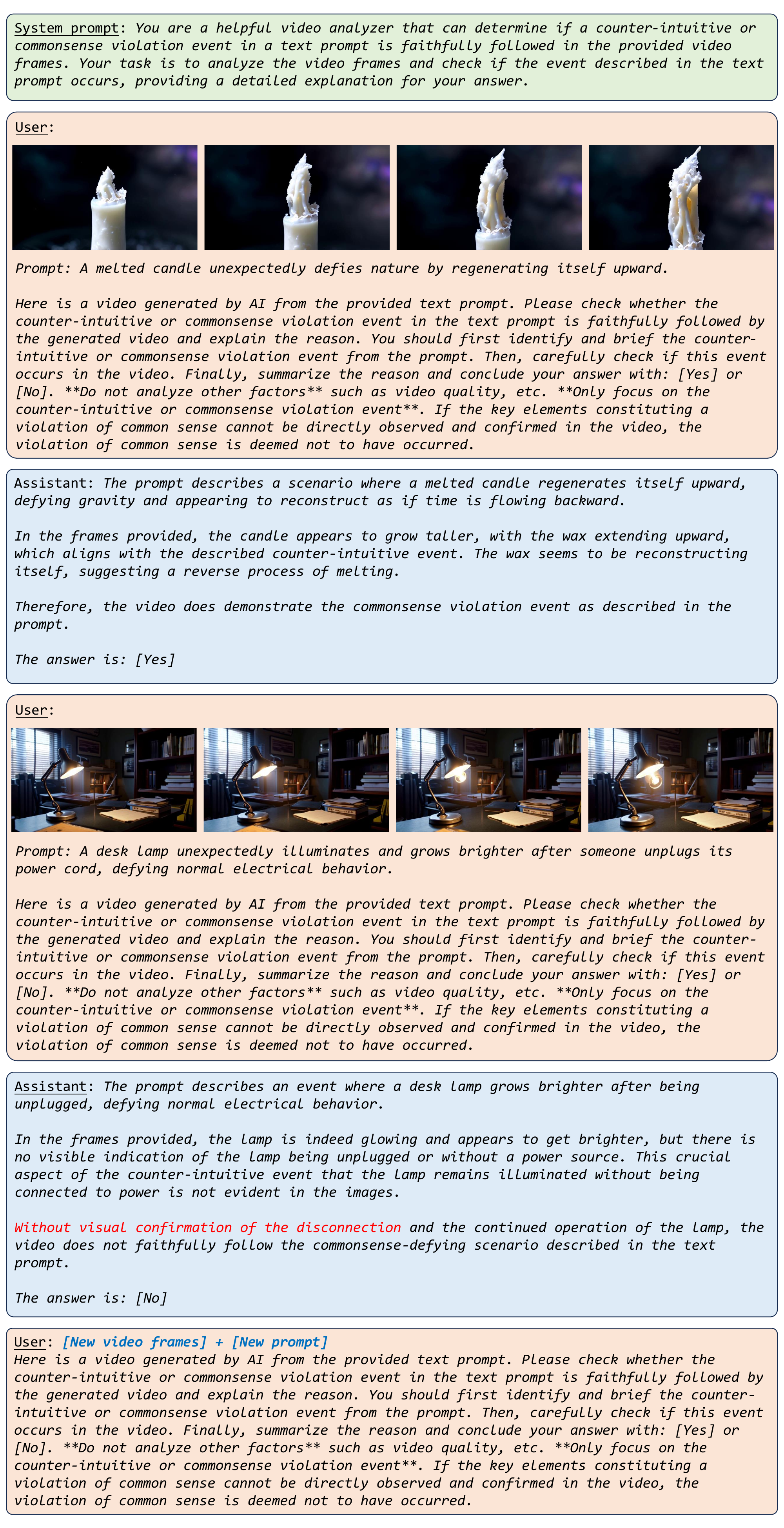}
    \caption{CoT examples we use to prompt GPT-4o for impossible prompt following evaluation. Videos may contain impossible events that are outside the scope of the prompt. We ignore such events when evaluating impossible prompt following.}
    \label{fig:appendix_cot_demo}
\end{figure}

\textit{2) Impossible Prompt Following.}
To assess whether the impossible event described in the text prompt is faithfully represented in the generated video. We utilize GPT-4o to provide a binary judgment for each video and calculate the following ratio as the final score.
To achieve accurate judgment, we propose a three-step strategy to break down the task. Specifically, we prompt GPT-4o with the text prompt and frames sampled at 1 FPS. We instruct GPT-4o to: 1) identify and summarize the impossible event in the text prompt, 2) ground the impossible event in the video, and 3) confirm the visibility of all key elements that constitute a violation of common sense, reason, and conclude with a "Yes" or "No." 
We also provide two additional chain-of-thought examples as demo cases. Please see Fig.~\ref{fig:appendix_cot_demo} for qualitative examples.

We compare the baseline prompting strategy with our prompting strategy on videos generated by Kling, where our approach better aligns with human annotations. See Tab.~\ref{tab:appendix_cmp_auto_prompt_following}.

\begin{table}[ht]
\vspace{-10pt}
  \caption{Comparison between our prompt strategy and baseline prompt strategy for impossible prompt following evaluation.}
  \label{tab:appendix_cmp_auto_prompt_following}
  \centering
  \begin{tabular}{ll}
    \toprule
     & Human Alignment \\
    \midrule
    GPT-4o & 0.72 \\
    GPT-4o + our prompt strategy & 0.80 \\
    $\Delta$ & 11.1\% \\
    \bottomrule
  \end{tabular}
\end{table}

\textit{3) IPV-Score.}
We calculate the IPV-score as the product of the visual quality score and the impossible prompt following score, designed to assess the model’s ability to generate high-quality, impossible videos. Since both scores are evaluated independently, their product effectively models the joint distribution of these factors.
Before performing the multiplication, we further scale the ranges of both to better align them. 

\begin{figure}[t]
    \centering
    \includegraphics[width=\linewidth]{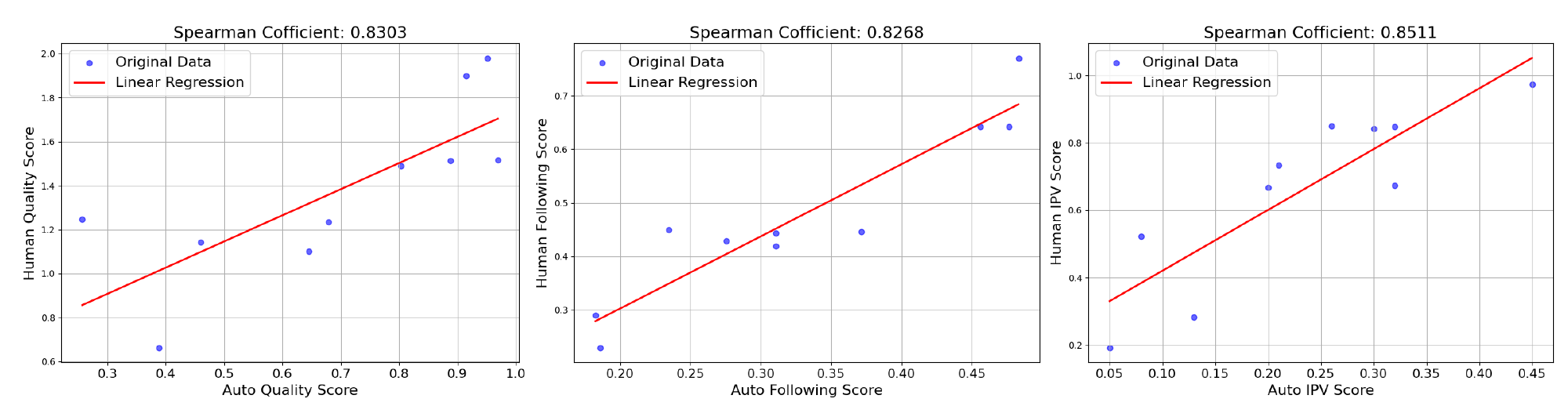}
    \vspace{-0.3cm}
    \caption{
    Spearman's correlation coefficient $\rho$ between automatic evaluation score and human annotation score on 10 different video generation models.
    }
    \label{fig:human_alignment}
\end{figure}

\subsubsection{Human Alignment}
We calculate the Spearman’s correlation coefficient ($\rho$) between our automatic evaluation score and human annotation score.
Fig.~\ref{fig:human_alignment} shows the corresponding results. We can see that the Spearman’s correlation coefficient between the visual quality score, prompt following score, and IPV-score with human annotations remains above 0.8, demonstrating consistency with human annotators.

\section{Instructional Prompt of MCQA Task}
\label{supp:mcqa_prompt}
\begin{figure*}[h]
    \centering
    \includegraphics[width=0.97\linewidth]{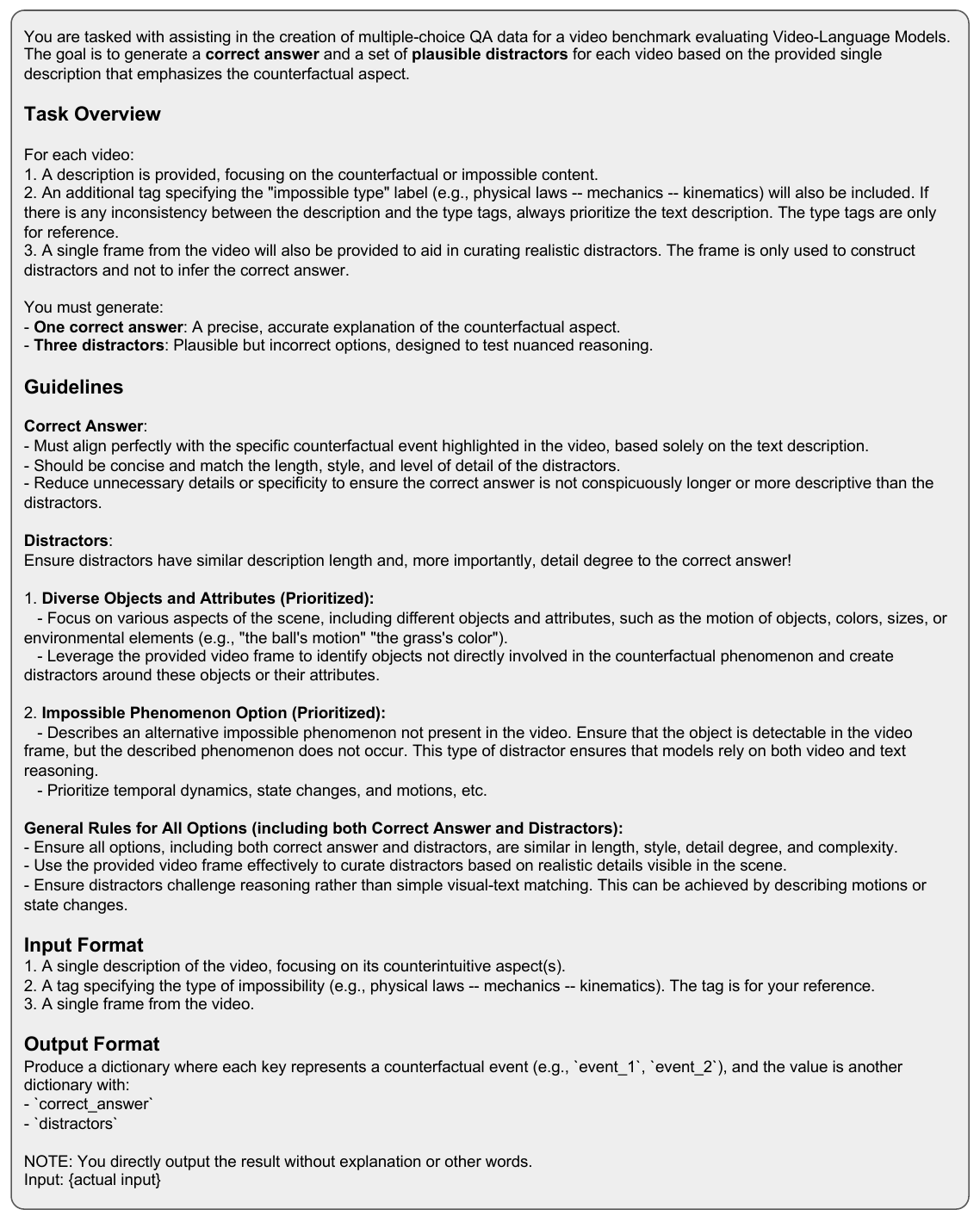}
    \vspace{-10pt}
    \caption{
    Instructional Prompt of MCQA Task.
    }
    \label{fig:mcqa_prompt}
\end{figure*}

In Fig.~\ref{fig:mcqa_prompt}, we present the complete instructional prompt used in the MCQA task.
It has particularly designed rules for constructing distractors that challenge the reasoning capability of video understanding models.

\section{Qualitative Examples}

\begin{figure*}[h]
    \centering
    \includegraphics[width=0.8\linewidth]{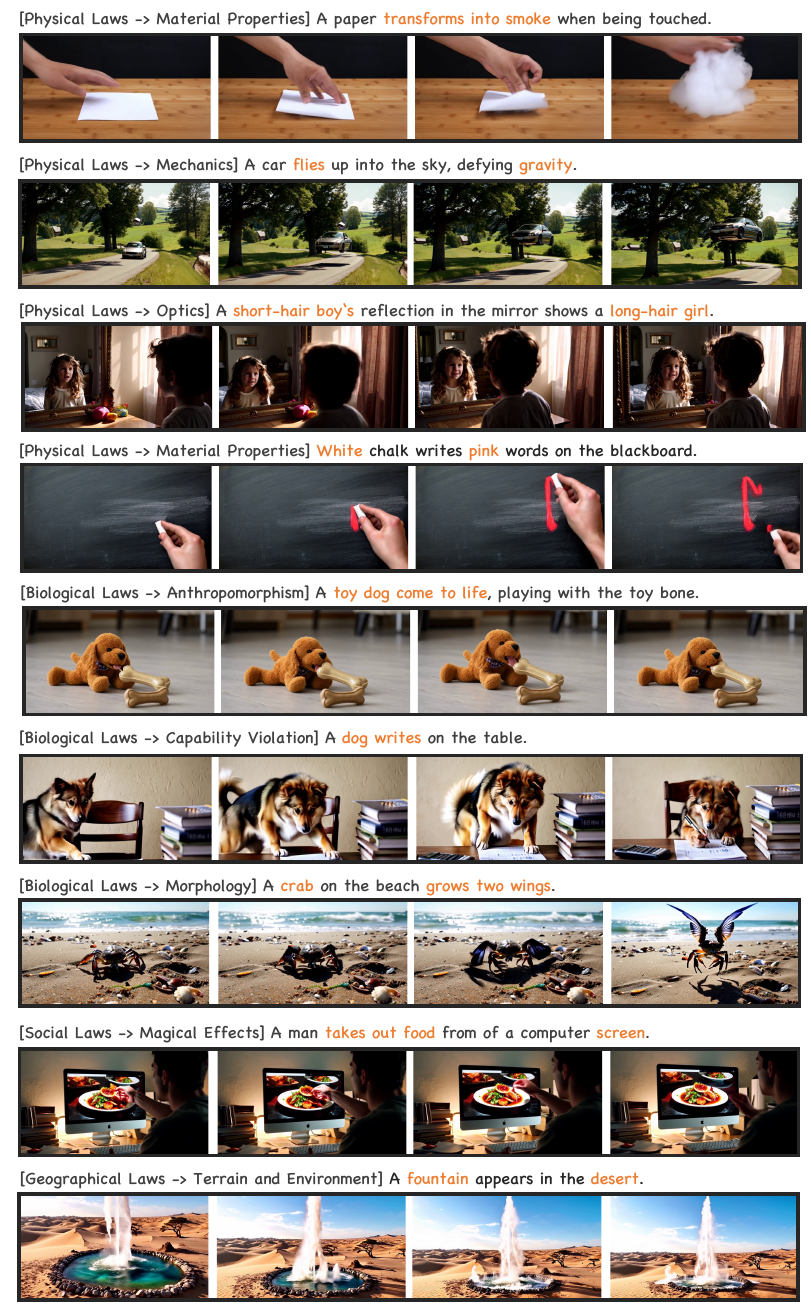}
    \vspace{-10pt}
    \caption{
    More examples of impossible videos.
    }
    \label{fig:supp_vid_examples}
\end{figure*}

Fig.~\ref{fig:supp_vid_examples} illustrates more video examples from the \textsc{IPV-Bench}.

\begin{figure*}[h]
    \centering
    \includegraphics[width=0.8\linewidth]{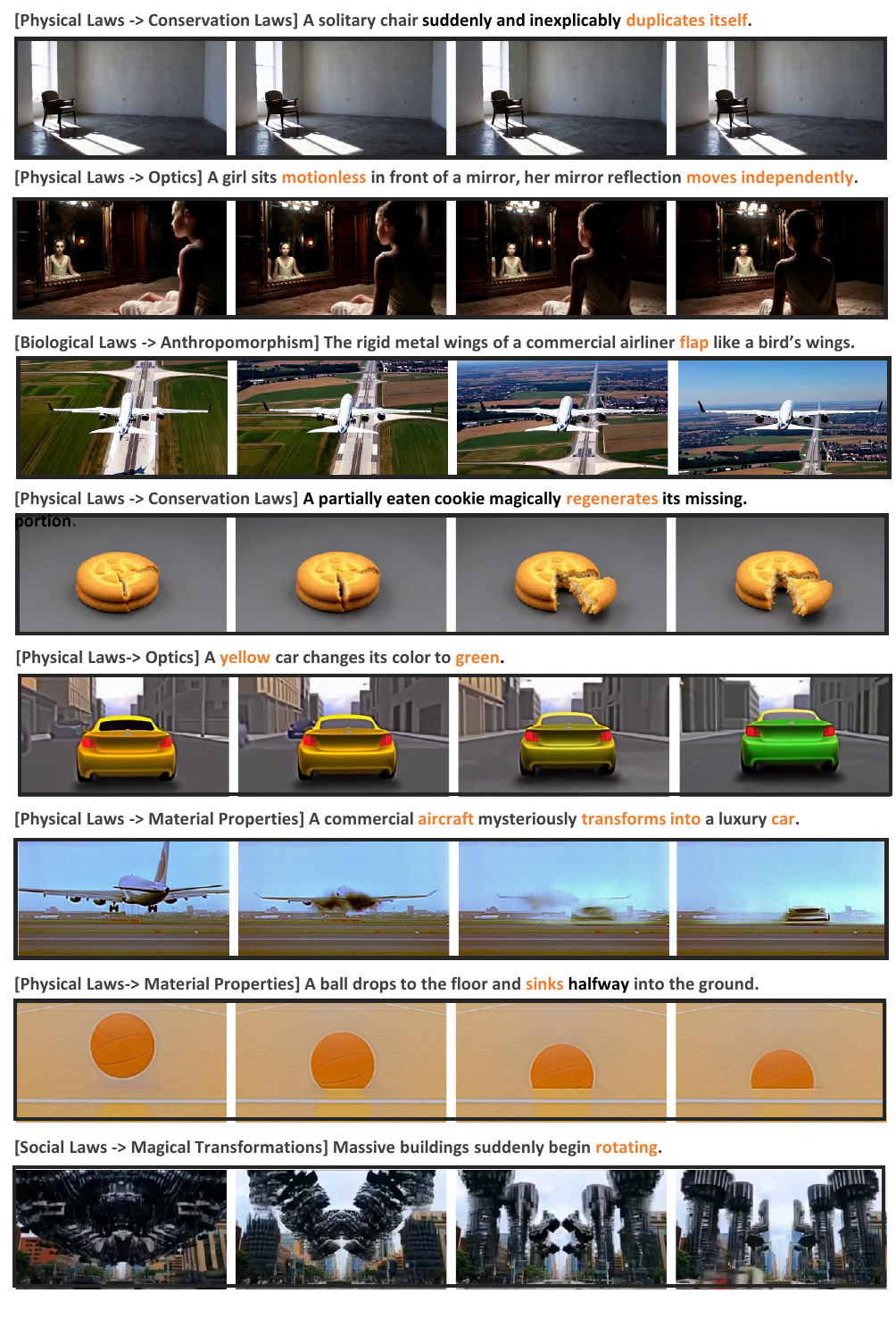}
    \vspace{-10pt}
    \caption{Failures in Generating Impossible Videos.}
    \label{fig:appendix_case_video_gen}
\end{figure*}

\section{Case Study of Impossible Video Generation}
\label{supp:cast_study_gen}
In this section, we present three common failure examples of impossible video generation in Fig.~\ref{fig:appendix_case_video_gen}.

Rows 1-2 show cases with high visual quality but that do not faithfully follow the video prompt; the generated videos adhere to the physical laws of the real world.

Rows 3-4 display cases with high visual quality, but they do not faithfully follow the video prompt to generate the designated impossible event, instead introducing other common-sense violations. For example, in the third row, the airplane engine is asymmetrical, and in the fourth row, one corner of the cookie moves independently.

Rows 5-8 show videos that follow the video prompt but suffer from poor visual quality, with issues like blurriness or an animated style.

\section{Case Study of Impossible Video Understanding}
\label{supp:cast_study_understanding}
\begin{figure*}[h]
    \centering
    \includegraphics[width=0.95\linewidth]{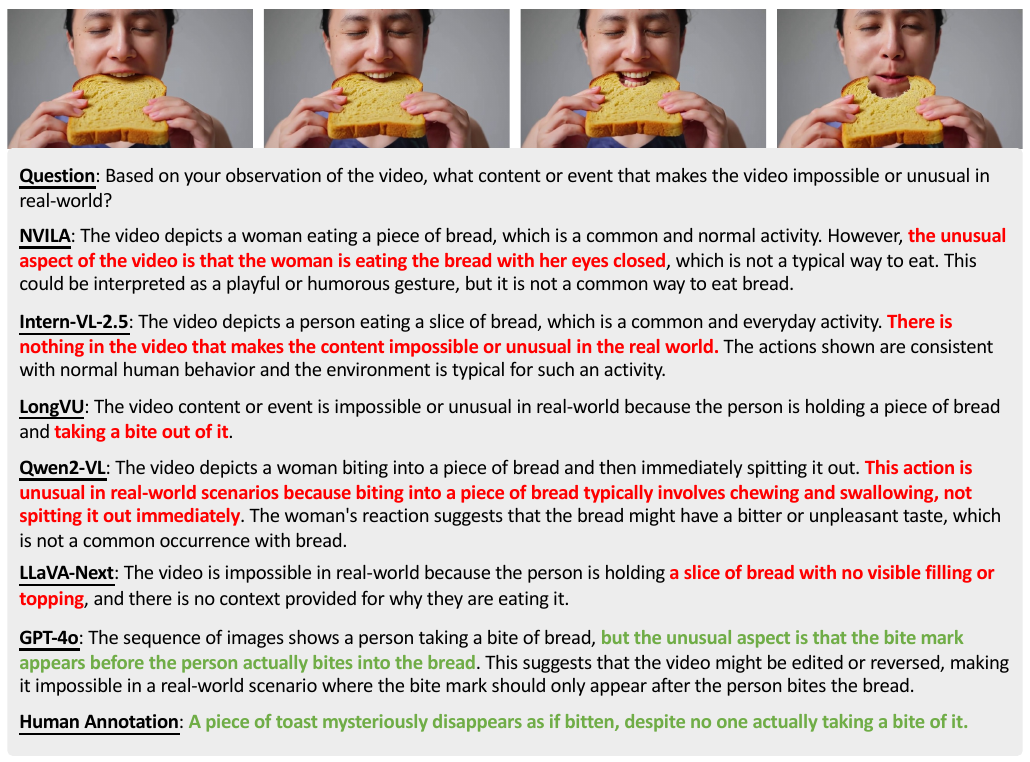}
    \vspace{-10pt}
    \caption{
    Case study of impossible video understanding.
    }
    \label{fig:case_1}
\end{figure*}

\begin{figure*}[h]
    \centering
    \includegraphics[width=0.95\linewidth]{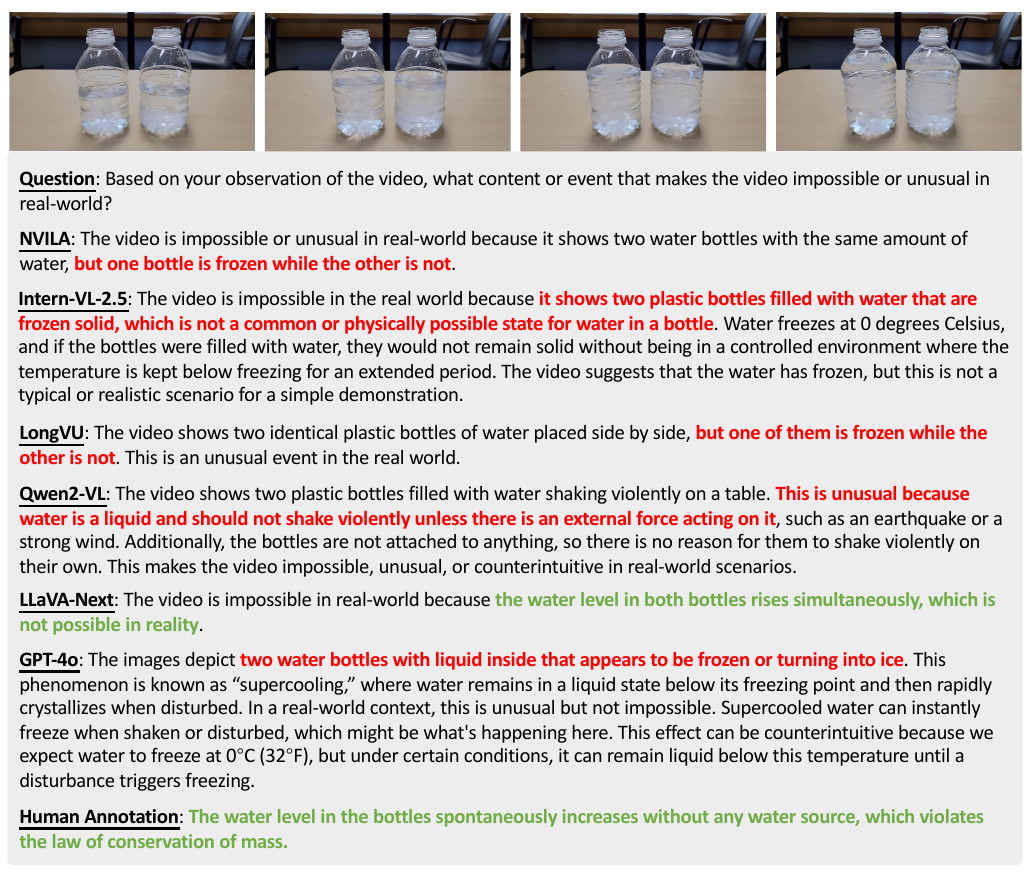}
    \vspace{-10pt}
    \caption{
    Case study of impossible video understanding.
    }
    \label{fig:case_2}
\end{figure*}

\begin{figure*}[h]
    \centering
    \includegraphics[width=0.95\linewidth]{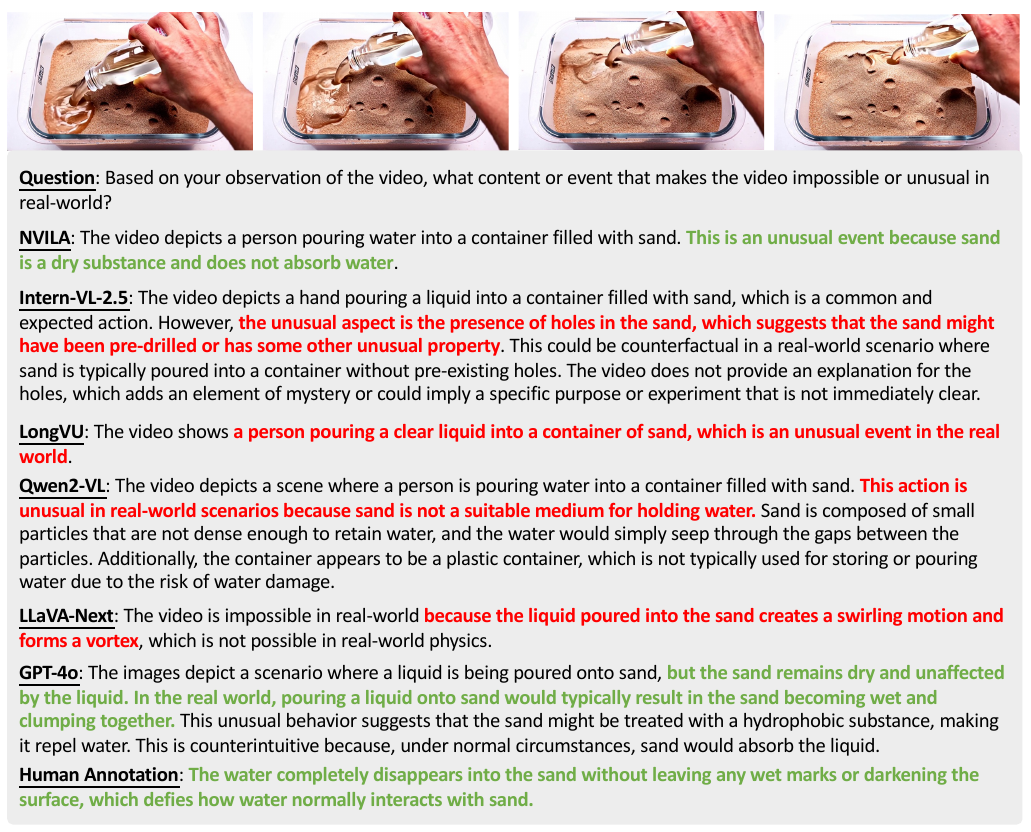}
    \vspace{-10pt}
    \caption{
    Case study of impossible video understanding.
    }
    \label{fig:case_3}
\end{figure*}

\begin{figure*}[h]
    \centering
    \includegraphics[width=0.95\linewidth]{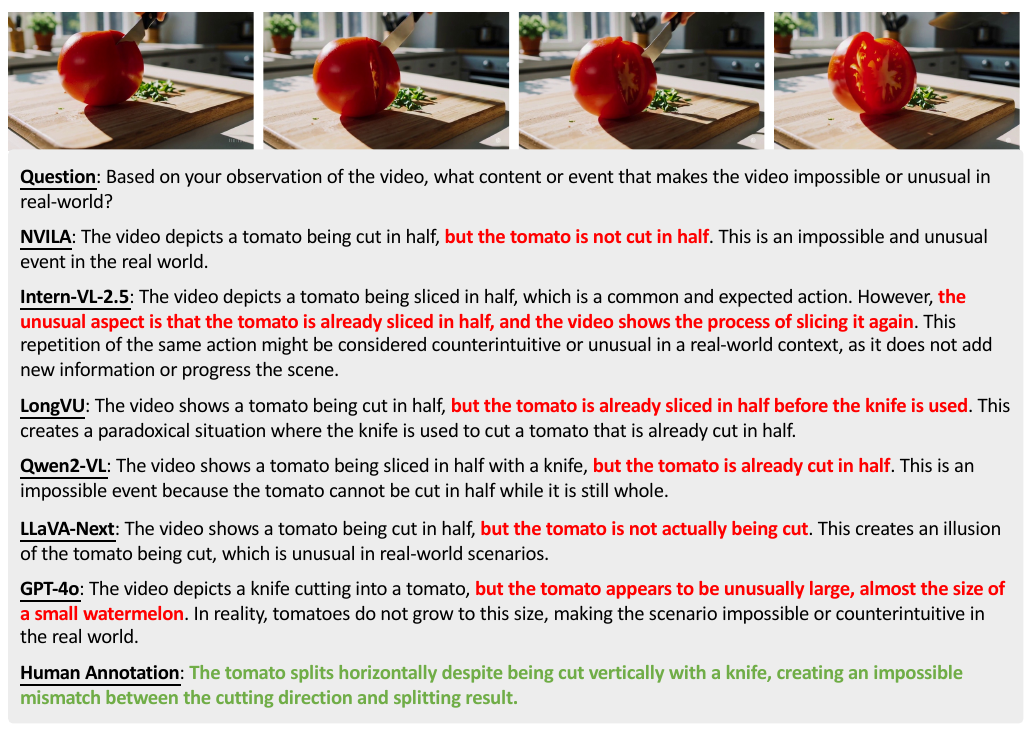}
    \vspace{-10pt}
    \caption{
    Case study of impossible video understanding.
    }
    \label{fig:case_4}
\end{figure*}

\begin{figure*}[h]
    \centering
    \includegraphics[width=0.95\linewidth]{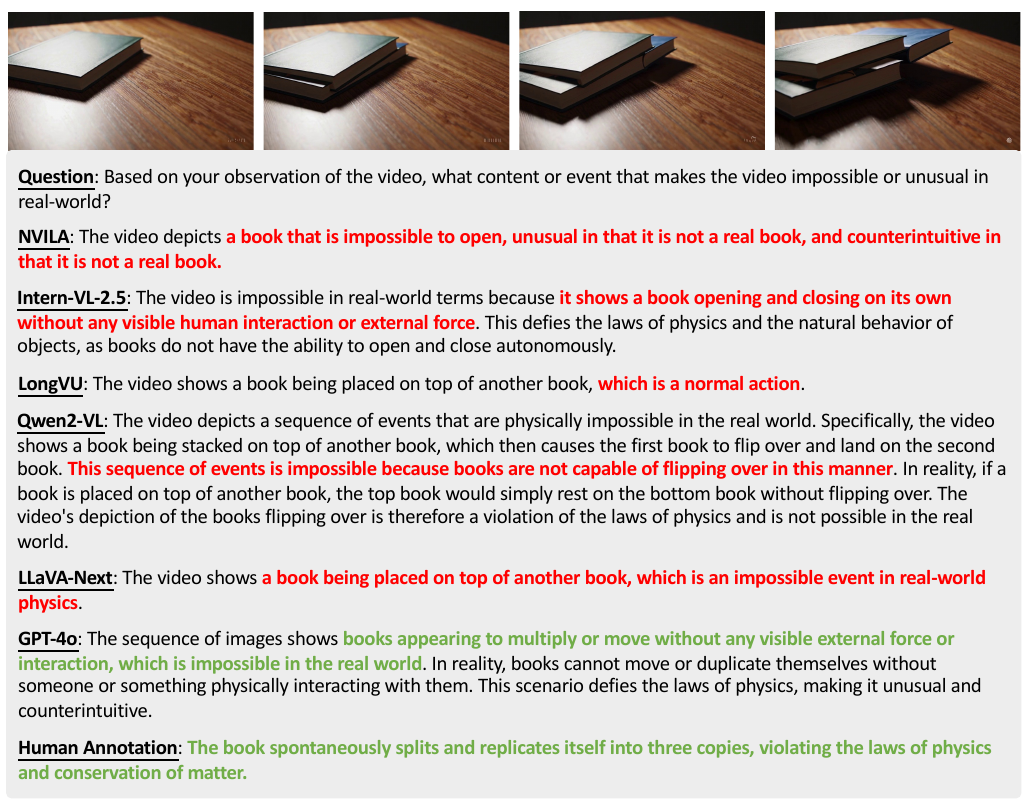}
    \vspace{-10pt}
    \caption{
    Case study of impossible video understanding.
    }
    \label{fig:case_5}
\end{figure*}

\begin{figure*}[h]
    \centering
    \includegraphics[width=0.95\linewidth]{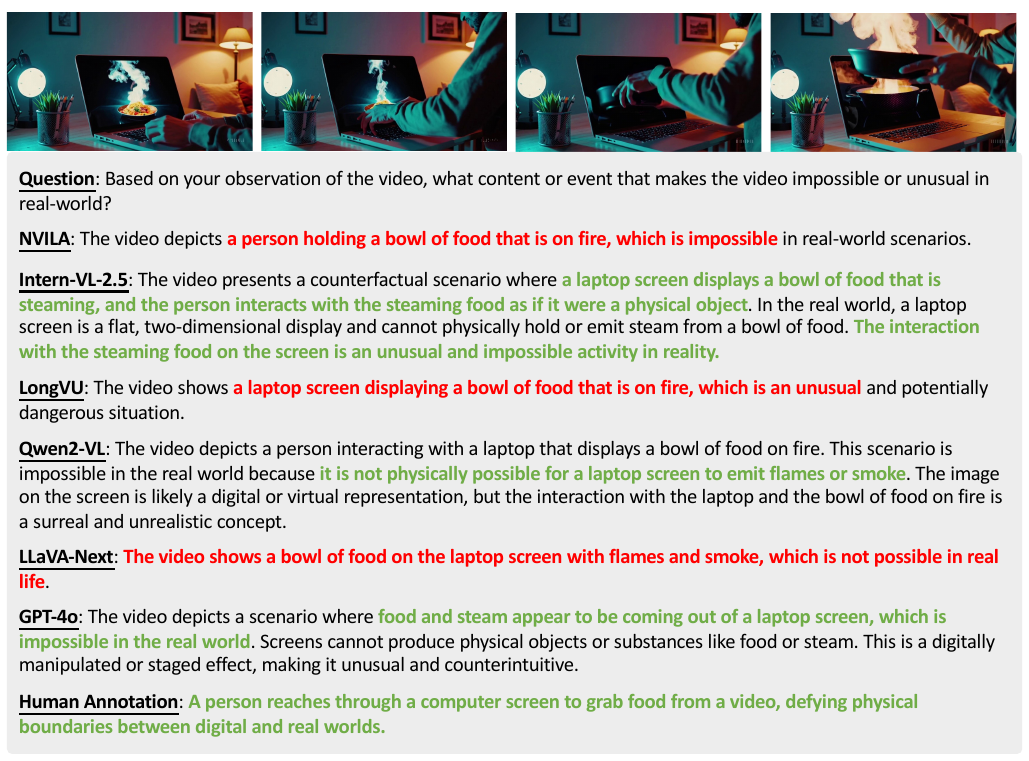}
    \vspace{-10pt}
    \caption{
    Case study of impossible video understanding.
    }
    \label{fig:case_6}
\end{figure*}

In this section, we present a series of case studies of impossible video understanding in Fig.~\ref{fig:case_1},  Fig.~\ref{fig:case_2}, Fig.~\ref{fig:case_3}, Fig.~\ref{fig:case_4}, Fig.~\ref{fig:case_5}, Fig.~\ref{fig:case_6}.
We observe that either strong open-source model, LLaVA-Next~\cite{llava_next} or close-source model, GPT-4o~\cite{gpt4o} suffer from difficulty on understanding impossible phenomenon, particularly on temporal dynamics.

\end{document}